\documentclass[pdflatex,sn-nature,Numbered]{sn-jnl}

\usepackage{graphicx}
\graphicspath{{Figure/}}
\usepackage{multirow}
\usepackage{amsmath,amssymb,amsfonts}
\usepackage{booktabs}
\usepackage{xcolor}
\usepackage{textcomp}
\usepackage{hyperref}
\usepackage{etoolbox}

\raggedright

\makeatletter
\renewcommand{\equalcont}[1]{\gdef\@equalconttext{#1}\g@addto@macro\artauthors{$^{,*}$}\global\equalconttrue}
\renewcommand{\@@corrauthor}[2][]{\def\@authfrstarg{#1}\@corauemailtrue\advance\corraucount by 1%
\g@addto@macro\artauthors{%
    \global\@auemailtrue%
    \Authorfont%
    \def\baselinestretch{1}%
    \authorsep{#2}\unskip%
    \ifx\@authfrstarg\empty%
      \textsuperscript{\textdagger}\hskip-1pt%
    \else%
      \textsuperscript{\smash{{%
        \@for\@@affmark:=#1\do{\edef\affnum{\@ifundefined{X@\@@affmark}{\@@affmark}{\jmkRef{\@@affmark}}}%
        \unskip\sep\affnum\let\sep=,}%
      }},\textdagger}\hskip-1pt%
    \fi\unskip%
    \def\authorsep{\au@and~}%
    \global\let\sep\@empty\global\let\@corref\@empty%
}}%
\patchcmd{\@maketitle}{\ifnum\aucount>1*\fi}{\ifnum\aucount>1$^{\dagger}$\fi}{}{}
\patchcmd{\@maketitle}{\par$^{\dagger}$\@equalconttext\par}{\par$^{*}$\@equalconttext\par}{}{}
\makeatother

\begin{document}

\title[Mapping a Collective Capability Boundary in Clinical LLMs]{A collective capability boundary in frontier large language models on guideline-conformant and case-specific oncology decision-making}

\author[1,6]{\fnm{Sheng} \sur{Zhang}}\email{dcszhang@foxmail.com}
\equalcont{These authors contributed equally to this work.}

\author[2,3]{\fnm{Jinming} \sur{Li}}\email{jmli16@fudan.edu.cn}
\equalcont{These authors contributed equally to this work.}

\author[4]{\fnm{Wangyang} \sur{Chen}}\email{chenwangyang99@163.com}
\equalcont{These authors contributed equally to this work.}

\author*[5]{\fnm{Zhiwei} \sur{Bao}}\email{zwbao1996@zju.edu.cn}

\author*[1]{\fnm{YoSean} \sur{Wang}}\email{yosean4desci@gmail.com}

\affil[1]{\orgdiv{Institute of Digital Medicine}, \orgname{City University of Hong Kong}, \orgaddress{\city{Hong Kong SAR}, \country{China}}}

\affil[2]{\orgdiv{Department of Medical Oncology}, \orgname{Fudan University Shanghai Cancer Center}, \orgaddress{\city{Shanghai}, \country{China}}}

\affil[3]{\orgdiv{Department of Colorectal Surgery}, \orgname{Fudan University Shanghai Cancer Center}, \orgaddress{\city{Shanghai}, \country{China}}}

\affil[4]{\orgdiv{Department of Gastroenterology}, \orgname{Affiliated Hangzhou First People's Hospital, Westlake University School of Medicine}, \orgaddress{\city{Hangzhou}, \state{Zhejiang}, \country{China}}}

\affil[5]{\orgdiv{College of Biomedical Engineering and Instrument Science, Ministry of Education Key Laboratory of Biomedical Engineering}, \orgname{Zhejiang University}, \orgaddress{\city{Hangzhou}, \country{China}}}

\affil[6]{\orgname{Shaoxing Vocational \& Technical College}, \orgaddress{\city{Shaoxing}, \state{Zhejiang}, \country{China}}}

\abstract{Large language models (LLMs) achieve impressive scores on medical knowledge examinations, yet real-world oncology is not a knowledge test---it is a sequence of guideline-pathway choices, escalation judgments, and commitments under uncertainty. Existing benchmarks largely measure factual recall on isolated questions, leaving open whether frontier LLMs share systematic blind spots in clinical decision-path reasoning and whether combining models can compensate. We built the Oncology Decision Boundary Benchmark (ODBB)---2{,}005 oncology decision points across NCCN guidelines and colorectal cancer cases---and evaluated nine frontier LLMs (four closed-source, five open-weight families) released between June 2025 and April 2026. A fully deterministic scorer (zero LLM inference) classified outputs into 14 failure types, independently validated by two oncologists (Cohen's weighted $\kappa$ = 0.939 and 0.790 for the two reviewers; 95\% bootstrap CIs [0.870, 0.985] and [0.671, 0.892]) on a 225-item stratified sample. Treating the nine as a pooled super-model, 42.1\% (Wilson 95\% CI 40.0\%--44.3\%) of all items---35.7\% of the 1{,}586 NCCN guideline items and 66.4\% of the 419 colorectal cancer case items---were answered correctly by none---failures concentrated in tasks requiring choice between guideline pathways before reasoning within any, indicating a consistent blind spot in clinical \emph{meta}-judgment across this cohort, likely to require architectural intervention rather than additional training data. Two models tuned for decisiveness (GPT-5.5, Gemini 3.1 Pro Preview) made unsafe commitments between three and five times more often than the seven cautious models without scoring higher overall---in our sample, agentic tuning did not consistently yield net improvement. Between 3\% and 9\% of items, models articulated the correct next clinical step in their reasoning yet did not commit to it as the final answer---failures of decision, not knowledge gaps. Model quality is no longer the primary bottleneck for clinical LLM deployment. The dominant constraint is now the prevailing assumption that any single model can serve as the sole basis for clinical decisions. Progress requires deployment architectures that detect when a model has reached its competence boundary and route the decision to a human clinician.
}


\maketitle

\section{Introduction}\label{sec:intro}

The rapid improvement of large language models on medical knowledge examinations has generated considerable optimism about their clinical utility. Models now routinely exceed passing thresholds on USMLE-style assessments and achieve high accuracy on curated question banks such as MedQA, MultiMedQA, and differential-diagnosis benchmarks~\cite{mcduff2023medbench, singhal2023medpalm, singhal2025medpalm2, nori2023gpt4, kung2023chatgpt, jin2021medqa, jin2019pubmedqa, hendrycks2021mmlu, lievin2024medqa}. These results have been widely interpreted as evidence that LLMs are approaching---or have already reached---clinical-grade reasoning capability~\cite{thirunavukarasu2023llmmedicine, clusmann2023llmclinical, saab2024medgemini}.

This interpretation rests on an implicit assumption: that high scores on knowledge-retrieval benchmarks translate to reliable performance on the decision-path navigation tasks that constitute actual clinical practice~\cite{topol2019highperformance}. In oncology, a treating physician does not merely recall facts; they determine which guideline pathway applies, identify what information is missing before a decision can be made, recognize when a patient should be routed to an alternative management track, and distinguish situations that permit action from those that demand restraint. These tasks require not only domain knowledge but also meta-decision competence---the ability to judge which reasoning framework to apply before reasoning within it~\cite{croskerry2009clinical, kahneman2011thinking}.

Existing clinical LLM evaluations have three notable gaps. First, most benchmarks assess factual recall rather than decision-path navigation, and the few that attempt clinical scenarios typically rely on open-ended vignettes scored by LLM-as-judge pipelines, introducing reproducibility concerns and self-evaluation bias~\cite{zheng2023judging, agrawal2025evaluation, chang2024survey}. Second, safety-relevant failure modes---premature commitment to treatment, contraindicated recommendations, failure to request missing information---are rarely disaggregated; a model that confidently gives a wrong but plausible answer receives the same ``incorrect'' label as one that commits a clinically dangerous error~\cite{ji2023hallucination, pal2023medhalt}. Third, no prior work has systematically characterized whether frontier LLMs share \emph{collective} blind spots in clinical reasoning: areas where the entire model class fails, and where adding more models to an ensemble yields negligible improvement.

In this study, we ask a different question from the conventional ``which LLM performs best.'' We ask: \emph{where is the collective capability boundary of current frontier LLMs in clinical decision-making, what is this boundary made of, and can model diversity breach it?}

To answer this, we constructed the Oncology Decision Boundary Benchmark (ODBB), comprising 2{,}005 clinical decision points across two complementary tracks: 1{,}586 structured guideline-navigation items derived from NCCN oncology guidelines, and 419 case-specific therapy items derived from published colorectal cancer case reports. All items are scored by a fully deterministic evaluation pipeline with 14 typed failure labels. We evaluated nine frontier LLMs spanning four closed-source and five open-weight model families. Our contributions are threefold:

\begin{enumerate}
    \item \textbf{An automatically regenerable oncology decision benchmark.} ODBB ships not only as a 2{,}005-item evaluation dataset, but as an open-source generation pipeline that ingests NCCN guideline workspaces and produces structured decision items across six clinically motivated question types. When NCCN releases a new guideline edition, the community can regenerate a fresh benchmark in a single run, avoiding the staleness that has shortened the useful lifespan of prior clinical LLM evaluations.
    \item \textbf{A deterministic, clinician-validated scorer.} We built a fully deterministic scoring engine (zero LLM inference) with 14 failure-type labels, then validated its judgments against two independent oncologists on a 225-item stratified sample (weighted $\kappa = 0.790$--$0.939$). Unlike LLM-as-judge pipelines, the scorer is bit-for-bit reproducible, eliminates self-evaluation bias, and now carries empirical reliability evidence from practicing clinicians---not just internal consistency.
    \item \textbf{Four findings that reframe how to read frontier LLM evaluations.} (a) Pooling all nine models still leaves 42\% of items unsolved, indicating a consistent capability boundary across this cohort that adding more models from these families cannot breach. (b) These failures concentrate in tasks requiring choice between guideline pathways before reasoning within any---a blind spot in clinical \emph{meta}-judgment, not knowledge. (c) Models tuned for decisiveness make unsafe commitments three to five times more often without scoring higher overall, contradicting the assumption that agentic behavior is a free upgrade. (d) Across 3\% to 9\% of items, models articulate the correct next step in their reasoning yet fail to commit to it---failures of decision rather than knowledge gaps.
\end{enumerate}

\section{Methods}\label{sec:methods}

\subsection{Study design overview}\label{sec:study_design}

This is a benchmark-introduction study evaluating nine frontier large language models on oncology clinical decision-making. The methodology comprises five components, described in turn in the subsections that follow: (i) the Oncology Decision Boundary Benchmark (ODBB) itself (\S\ref{sec:benchmark}), 2{,}005 scorable items across two complementary tracks---NCCN structured decision items and CRC case-based therapy recommendation---generated through an automated, fully reproducible pipeline; (ii) a fully deterministic scorer with 14 typed failure labels (\S\ref{sec:scorer}), designed to eliminate the self-evaluation circularity inherent to LLM-as-judge pipelines; (iii) nine evaluated frontier LLMs covering closed-source and open-weight model families released December 2025--April 2026 (\S\ref{sec:models}); (iv) a two-wave clinician adjudication by board-certified oncologists (\S\ref{sec:clinician}), providing reliability evidence for both scorer verdicts and gold-answer validity on the items underpinning the headline capability boundary; and (v) statistical analysis (\S\ref{sec:stats}) using bootstrap confidence intervals, permutation tests for tier separation, and Wilson score intervals for proportions. All datasets, scorer source, raw model outputs, and adjudication data are openly released under the manuscript's Data Availability and Code Availability statements.

\subsection{Benchmark composition}\label{sec:benchmark}

\paragraph{Overview.} ODBB comprises two complementary item tracks totaling 2{,}005 scorable clinical decision points. The first track consists of 1{,}586 structured decision items derived from 69 NCCN cancer-type guideline workspaces (snapshot frozen as of March 2026)~\cite{nccn2025guidelines}, covering the full scope of solid and hematologic malignancy management. Each item presents a clinical scenario at a specific decision node within a guideline pathway and requires the model to determine the appropriate next management step. The second track consists of 419 colorectal cancer (CRC) case-based items derived from published case reports, requiring specific therapy recommendations evaluated against structured gold-standard treatment regimens. All gold-standard answers and item content are SHA-256 hash-locked for reproducibility.

\paragraph{Automated generation pipeline.} Items are produced by an eight-stage automated pipeline that ingests a raw NCCN PDF and emits structured, schema-validated decision items. Five stages use a single auxiliary LLM (Doubao Seed 2.0 Pro, ByteDance Seed)~\cite{doubaoSeed2pro} for semantic judgment (per-page classification, anchor--evidence link judgment, local subgraph drafting, cancer-specific slot discovery, item strategy authoring); three stages perform deterministic structural assembly (anchor pooling and tree-seed materialization, global graph merging with four typed registry construction, structured item-JSON generation with clinical consistency audit). All stages emit SHA-256 hash-locked artifacts with explicit invalidation matrices, and NCCN guideline updates trigger full pipeline re-run without manual authoring. The pipeline architecture is depicted in Figure~\ref{fig:pipeline} (Section~\ref{sec:benchmark_overview}), where individual stage roles are described.

\paragraph{Question type design.} Items span six clinically motivated question types that map to distinct decision flavors encountered in oncology practice. \emph{Parallel option disambiguation} ($n = 920$) is the modal task: at a guideline node where two or more permitted treatment options coexist, the model must select the option appropriate for the patient's specific situation (e.g., at a stage III colon cancer adjuvant chemotherapy node, NCCN permits both FOLFOX and CAPOX; selecting between them depends on patient comorbidity, oxaliplatin tolerance, and treatment-duration preferences). \emph{Missing information request} ($n = 348$) items require the model to recognize that available patient information is insufficient to commit to a decision and to request the specific missing variable rather than guess (e.g., a metastatic colorectal cancer patient with unspecified RAS/BRAF status requires molecular profiling before EGFR-targeted recommendation). \emph{In-guide handoff resolution} ($n = 136$) items require the model to route the patient to a different management section within the same NCCN guideline (e.g., a breast cancer patient developing brain metastases must be routed from systemic treatment to CNS metastasis management). \emph{Upstream routing decision} ($n = 110$) items require the model to select the correct guideline pathway \emph{before} reasoning within any one of them; this is the form of meta-judgment that frontier LLMs most consistently fail on (Section~\ref{sec:wall}; e.g., a hepatocellular carcinoma patient with compensated cirrhosis must first be routed to the resection / transplant / systemic-therapy branch based on Barcelona Clinic Liver Cancer staging). \emph{Evidence resolution request} ($n = 69$) items present a clinical situation consistent with multiple guideline pathways, requiring the model to identify the specific additional evidence needed to disambiguate (e.g., a melanoma patient with positive sentinel lymph node biopsy who could enter either observation or adjuvant immunotherapy must specify the evidence needed to choose). \emph{Single-path decision} ($n = 3$) items are specialized cases where only one management pathway is supported by the guideline at the given node; included for completeness. Question types are determined deterministically from the underlying pathway-graph topology and slot resolution status, not by free-form authoring.

\paragraph{Rationale for two complementary tracks.} The two tracks correspond to two qualitatively distinct cognitive tasks: the NCCN track resembles a specialty board examination (single-step decisions against a \emph{normative} gold standard---what the guideline says one should do), while the CRC case track resembles a case-conference discussion (holistic reasoning against a \emph{descriptive} gold standard---what the treating physician actually did). Combining both in a single benchmark stress-tests LLMs along two orthogonal axes---``mastery of the guideline framework'' and ``ability to reason over real case narratives''---under the premise that adequacy on one track alone does not establish clinical readiness. Concretely: the NCCN track tests guideline mastery at structured decision nodes (given the guideline-defined patient state, the model commits to the next pathway step), and the CRC case track tests free-text case interpretation (extracting key clinical facts---stage, molecular profile, prior treatment lines, comorbidities---from a narrative case report, integrating them, and recommending a specific therapy regimen). Beyond the mechanical question of whether a model can match either the guideline or the case-report regimen, the two tracks together probe a deeper question: \emph{when a model fails on consensus guideline conformity, does it retain any independent capacity for reasoning over atypical real-world cases reported in the literature?} The CRC track, by anchoring its gold standard to non-mainstream published management, is specifically designed as a probe for this case-specific reasoning capacity.

\paragraph{Rationale for CRC selection as the case-based track.} Colorectal cancer (CRC) was selected as the case-based evaluation track for four reasons. First, CRC has a mature and extensive published case-report literature, ensuring sufficient sample size for statistically meaningful evaluation. Second, modern CRC therapy is highly structured at the regimen level---specific cytotoxic backbones (FOLFOX, FOLFIRI, CAPOX), targeted agents (anti-EGFR, anti-VEGF, BRAF/MEK inhibitors), and immune checkpoint blockade---allowing canonical drug-list normalization at strict, class, and family granularity. Third, CRC treatment selection depends heavily on biomarker-driven branching (RAS, BRAF, MSI/MMR, HER2), providing the molecular-context reasoning complexity that ODBB's question types are designed to probe. Fourth, the authors' team includes colorectal-cancer subspecialists, enabling rigorous clinician audit of gold-standard regimens. Extension of the case-based framework to other malignancies (lung, breast, hematologic) is a deliberate future-work direction (Section~\ref{sec:limitations}).

\paragraph{PubMed search strategy.} CRC case reports were retrieved from PubMed (\url{https://pubmed.ncbi.nlm.nih.gov}) using the query \texttt{"colorectal cancer" AND "case report" AND (metastatic OR advanced OR recurrent)}, with publication date restricted to January 2020--April 2026 (plus a small number of older landmark cases retained for clinical-scenario diversity). The initial query yielded approximately 500 candidate articles. Manual screening against the inclusion criteria detailed below retained the final 419 items. The complete source index with PMIDs, journals, and publication dates is provided in Supplementary Material~S4.

\paragraph{CRC case sourcing.} The 419 CRC items were drawn from published case reports of metastatic and recurrent colorectal cancer (complete source index with PMIDs, journals, and publication dates in Supplementary Material~S4). Cases were included if (i) treatment course and clinical outcome were fully reported, (ii) staging, molecular profile (RAS / BRAF / MSI as reported per CAP/ASCP biomarker guidelines~\cite{sepulveda2017molecular}), and prior treatment lines were extractable, and (iii) the prescribed therapy regimen could be normalized to a structured drug list. Gold-standard regimens were canonicalized to three granularity levels---strict (exact drug match), class (drug-class match using a standard oncology pharmacology taxonomy~\cite{who2024atc} with custom drug-to-class mappings detailed in Supplementary Material~S5), and family (broader functional drug-family match)---enabling sensitivity analysis on scoring stringency.

\paragraph{CRC gold-answer framing.} For CRC case items, we designate as the gold answer the regimen \emph{actually administered} in the published case report---i.e., the gold answer is a paper-team construct derived from each case's documented management decision, not necessarily the regimen that NCCN would recommend for that clinical scenario. The two often overlap but are not equivalent: published case reports systematically over-represent atypical, refractory, or experimental management (cases following standard NCCN regimens have lower publication value and are under-represented in the case-report literature). Consequently, a CRC item judged \texttt{incorrect} indicates that the model failed to reproduce the treating physician's specific decision in that case, \emph{not} necessarily that the model fails to know NCCN colorectal-cancer guidance. To prevent these two distinct constructs from being conflated, the Results section reports NCCN-track and CRC-track findings separately throughout (see also the gold-answer audit in Supplementary~S7, where the CRC subsample exhibits markedly lower clinical-correctness and NCCN-traceability rates than the NCCN-structured subsample), and the Discussion treats CRC-track evidence as evidence about case-specific decision reproduction rather than as evidence about guideline knowledge.

\subsection{Deterministic scorer}\label{sec:scorer}

\paragraph{Overview.} The scoring engine is implemented entirely in Python with zero LLM inference at any stage of evaluation. The pipeline operates in three sequential phases---schema validation and repair, decision alignment, and content scoring---augmented by a CRC-specific therapy-regimen evaluation track and a fine-grained typed failure-label classifier. All phases are deterministic and bit-for-bit reproducible: the same model output yields the same verdict on every re-run.

\paragraph{Phase 1: Schema validation and repair.} Model outputs are first validated against a JSON schema specifying the required clinical-decision fields (\texttt{decision}, \texttt{next\_step}, \texttt{needs} or \texttt{evidence}, \texttt{basis}, \texttt{reasoning}, etc.). When models emit common but recoverable deviations---field-name typos (\texttt{need} for \texttt{needs}, \texttt{rational} for \texttt{rationale}), unwrapped \texttt{item\_id} containers, non-standard enumeration values, or NCCN category labels in legacy formats---the scorer applies a curated set of repair rules to canonicalize the output before judging it. Outputs that remain unparseable after repair are labeled \texttt{technical\_unparseable\_json}; outputs that parse but violate the schema after repair are labeled \texttt{technical\_schema\_violation}. Both technical labels propagate to the final verdict as \texttt{technical\_failure} (excluded from accuracy computation but tracked separately for model-robustness analysis).

\paragraph{Phase 2: Decision alignment.} The scorer extracts the model's structured clinical decision (e.g., \texttt{stop\_missing\_info}, \texttt{stop\_need\_evidence}, \texttt{proceed\_with\_treatment}, \texttt{route\_to\_alternative\_pathway}) and compares it against the gold-standard decision at the level of decision intent. A model that selects the correct intent direction (stop versus proceed) receives credit for decision alignment even if its content scoring is only partial; this separation is essential for diagnosing whether failures arise from the high-level go/no-go judgment or from the downstream content. Decision-alignment outcomes feed three structured fields---\texttt{stop\_decision}, \texttt{downstream\_commitment}, and \texttt{decision\_aligned}---that are jointly used in verdict assignment.

\paragraph{Phase 3: Content scoring.} For NCCN \emph{stop} items (where the gold standard requires the model to request additional information), content scoring uses slot-level matching: the model's enumerated \texttt{needs} are normalized via a synonym dictionary, mapped to the gold-standard slot set, and scored with precision, recall, and F1. Token frequency across the full NCCN gold corpus is used to compute inverse-document-frequency (IDF) weights, which down-weight common slot tokens (e.g., ``status'', ``information'') in the matching score. For NCCN \emph{proceed} items, content scoring uses path-level matching: the model's \texttt{next\_step} is compared against gold-standard pathway-step strings using a 0.35 token-overlap threshold, with polarity-conflict detection to flag contraindicated recommendations that superficially share vocabulary with the gold answer. Both tracks produce a continuous content score in $[0, 1]$ that is combined with decision alignment to assign the final verdict (correct / partial / incorrect; verdict thresholds and the full slot-matching algorithm in Supplementary~S3).

\paragraph{CRC therapy-regimen scoring.} For CRC case items, the scorer canonicalizes both the model's recommended regimen and the gold-standard regimen to a structured drug list, then computes therapy F1 at three granularity levels: \emph{strict} (exact drug-name match after alias normalization, e.g., \texttt{FOLFOX4} $\rightarrow$ \texttt{folfox}, \texttt{Lonsurf} $\rightarrow$ \texttt{ftd/tpi}); \emph{class} (drug-class match using a 43-entry custom dictionary aligned with standard oncology pharmacology nomenclature; see Supplementary~S5); and \emph{family} (broader functional drug-family match). A CRC verdict is \emph{correct} when strict therapy F1 $\geq 0.67$ \emph{and} treatment intent matches \emph{and} no contraindicated drugs are flagged; \emph{partial} when F1 $> 0$; \emph{incorrect} otherwise. The three-tier granularity supports sensitivity analysis: a model that recommends \texttt{capecitabine} instead of \texttt{5-FU} fails strict matching but passes class matching (both are fluoropyrimidines), making it possible to disentangle ``wrong drug'' from ``wrong drug class'' failures.

\paragraph{Typed failure-label taxonomy.} Beyond verdict assignment, the scorer assigns one or more of fourteen typed failure labels per item, capturing clinically distinct failure modes (Table~\ref{tab:failure_labels}). Labels of particular clinical significance include F1 (unsafe overreach: committing to action when the gold standard requires stopping for information), F3 (premature commitment: selecting a specific treatment before confirming eligibility), F5 (false stop: halting when sufficient information exists to proceed), F10 (evidence bypass: proceeding without resolving an evidence conflict the case explicitly flagged), and F16 (contraindicated recommendation: a regimen the gold standard explicitly rules out). Labels F2, F8, F9, and F15 are reserved for forthcoming failure modes not yet observed at sufficient frequency to warrant operational definition. Operational triggers, decision thresholds, and example failure quotations from real model outputs are provided in Supplementary~S3.

\begin{table}[!t]
\centering
\caption{Fourteen typed failure labels assigned by the deterministic scorer.}\label{tab:failure_labels}
\small
\begin{tabular}{@{}llp{8.5cm}@{}}
\toprule
\textbf{Label} & \textbf{Surface} & \textbf{Failure mode (concise)} \\
\midrule
F1  & NCCN  & Unsafe overreach---committing to action when gold requires stop \\
F3  & NCCN  & Premature commitment---selecting a treatment before confirming eligibility \\
F4  & NCCN  & Missed decisive info---failing to request a slot critical to branching \\
F5  & NCCN  & False stop---halting when sufficient information exists to proceed \\
F6  & NCCN  & Overcautious---paired with F5 (co-emitted by the current scorer; see Supplementary~S3.7 for trigger detail) \\
F7  & NCCN  & Wrong route---selecting an incorrect pathway branch \\
F10 & NCCN  & Evidence bypass---proceeding without resolving a flagged evidence conflict \\
F11 & CRC   & Unsupported recommendation---regimen not supported by case profile \\
F12 & CRC   & Treatment intent mismatch---e.g., adjuvant vs.\ palliative confusion \\
F13 & CRC   & Molecular-context error---ignoring RAS / BRAF / MSI status \\
F14 & CRC   & Therapy-set mismatch---wrong drug combination \\
F16 & CRC   & Contraindicated recommendation---regimen the gold explicitly rules out \\
\midrule
\multicolumn{3}{l}{\textit{Technical labels (excluded from clinical-failure counts):}} \\
T1  & both  & \texttt{technical\_schema\_violation}---output parses as JSON but violates the track-specific schema after repair \\
T2  & both  & \texttt{technical\_unparseable\_json}---output cannot be parsed as JSON \\
\bottomrule
\end{tabular}
\begin{flushleft}
\footnotesize Note: Operational triggers and example quotations in Supplementary~S3. Labels F2, F8, F9, F15 are reserved for forthcoming failure modes; current active set comprises 12 clinical + 2 technical labels.
\end{flushleft}
\end{table}

\paragraph{Why deterministic rather than LLM-as-judge.} Unlike LLM-as-judge pipelines~\cite{zheng2023judging} and holistic evaluation frameworks that rely on model self-assessment~\cite{bommasani2023helm, srivastava2023beyond}, our approach eliminates all stochastic components from evaluation. This delivers three practical advantages. First, \emph{exact reproducibility}: re-running the scorer on the same outputs yields bit-identical verdicts, enabling longitudinal studies and head-to-head comparisons across publications without judge drift. Second, \emph{no self-evaluation bias}: closed-source frontier LLMs are not used to judge themselves or their competitors, removing a known source of unfair advantage~\cite{zheng2023judging}. Third, \emph{auditable failure attribution}: every verdict can be traced to a specific schema field, slot match, or threshold, making disagreements with clinician judgment debuggable rather than opaque. The complete scorer source code and a SHA-256 hash-locked release identifier are provided for third-party verification (Supplementary~S6).

\subsection{Models evaluated}\label{sec:models}

We evaluated nine frontier LLMs released between June 2025 and April 2026 (Table~\ref{tab:models}); eight of the nine were released between December 2025 and April 2026, and Gemini 2.5 Pro (June 2025) is retained as the immediate predecessor of Gemini 3.1 Pro Preview to support the within-vendor regression analysis in Section~\ref{sec:tradeoff}. The selection prioritized diversity across model families, licensing regimes (five open-weight, four closed-source), and price points. All models were accessed through the OpenRouter API; the exact OpenRouter model identifiers used at evaluation time are listed in Supplementary~Table~S6.6~\cite{openrouter2026models}. Each model was queried with temperature fixed at 0 and maximum output tokens set to 8,192. Output truncation was monitored via the T1/T2 technical-failure labels (Section~\ref{sec:scorer}); truncation-induced parsing failures occurred in below 1\% of items for eight of nine models (Supplementary~S3). Every model received an identical system prompt and item prompt with no model-specific tuning; complete prompt templates are available in the released code repository (see Data Availability and Code Availability statements).

\begin{table}[ht]
\centering
\caption{Models evaluated in ODBB. Prices in USD per million tokens, rounded to two decimal places; exact OpenRouter prices (e.g., Qwen 3.6-Plus input \$0.325, DeepSeek V4 Pro input \$0.435) are recorded in the released cost-log files.}\label{tab:models}
\begin{tabular}{@{}llccc@{}}
\toprule
Model & Release & Open-weight & Input (\$) & Output (\$) \\
\midrule
Claude Sonnet 4.6 & Feb 2026 & No & 3.00 & 15.00 \\
GPT-5.5 & Apr 2026 & No & 5.00 & 30.00 \\
Gemini 2.5 Pro & Jun 2025 & No & 1.25 & 10.00 \\
Gemini 3.1 Pro Preview & Feb 2026 & No & 2.00 & 12.00 \\
Qwen 3.6-Plus & Apr 2026 & Yes & 0.33 & 1.95 \\
DeepSeek V4 Pro & Apr 2026 & Yes & 0.44 & 0.87 \\
GLM-5 & Feb 2026 & Yes & 0.72 & 2.30 \\
GLM-5.1 & Apr 2026 & Yes & 0.95 & 3.15 \\
Minimax M2.7 & Mar 2026 & Yes & 0.30 & 1.20 \\
\bottomrule
\end{tabular}
\begin{flushleft}
\small Note: Input and output prices are per million tokens as listed on the OpenRouter API platform.
\end{flushleft}
\end{table}

\subsection{Clinician adjudication}\label{sec:clinician}

\paragraph{Overview and design rationale.} To validate that the deterministic scorer produces clinically defensible verdicts, two board-certified oncologists independently reviewed a stratified sample of 225 unique items, with 25 items reviewed by both reviewers for inter-rater reliability (250 reviewer--item pairs in total). After excluding 5 pairs marked \texttt{unsure} (2 by Doctor~A, 3 by Doctor~B), 245 evaluable scorer--clinician pairs remained (covering 220 unique items). Because comprehensive clinician adjudication of the full 2{,}005-item benchmark would have required approximately 67 reviewer-hours per oncologist (at $\sim$2 min/item), we adopted a stratified sampling design covering the highest-impact adjudication strata.

\paragraph{Sampling strategy.} The 225-item sample was drawn from three sources stratified by question type and scorer verdict. (i) An NCCN question-type stratum of 135 items allocated proportionally across the six question types (77 parallel option, 30 missing information, 11 in-guide handoff, 9 upstream routing, 5 evidence resolution, 3 single-path), preserving the natural distribution while guaranteeing minimum cell counts for inter-stratum comparison. (ii) A non-gastrointestinal workspace stratum of 60 items weighted across the 59 non-GI NCCN guideline workspaces (e.g., breast, lung, lymphoma, melanoma), avoiding overrepresentation of GI cancers that would otherwise dominate the structured item pool. (iii) A CRC case stratum of 30 items randomly sampled from the 419 CRC items. Items were further balanced across the nine evaluated models (each model receiving 32--33 of the 225 items). All sampling used a frozen random seed (recorded in Supplementary~S1) to ensure exact reproducibility.

\paragraph{Reviewer assignment.} Two board-certified oncologists each reviewed 125 items, with 25 items reviewed by both for inter-rater reliability estimation ($125 + 125 - 25 = 225$ unique items). Reviewer assignment within each stratum was randomized; the 25 overlap items preserved the same stratification structure (13 parallel option, 6 missing information, 2 in-guide handoff, 2 upstream routing, 1 evidence resolution, 1 single-path).

\paragraph{Review packet structure.} Each adjudicated item was presented as a structured record containing the item identifier, question type and source NCCN workspace, the structured clinical scenario (patient facts), the NCCN gold-standard answer (provided as reference), the LLM's structured output (decision, recommended next step, slots, basis, reasoning), and the deterministic scorer's verdict (correct / partial / incorrect) with its associated metrics (slot F1, content score, failure-label assignments). Clinicians were instructed to adjudicate whether the scorer's verdict was clinically appropriate given the LLM output and the NCCN gold standard---not to re-evaluate the NCCN gold-standard content itself. Even when both the scorer and the clinician reference the same gold standard, the clinician retains the independent ability to flag verdicts that mischaracterize the LLM's actual reasoning---the gold standard functions as a shared reference anchor for adjudication, not as a basis for tautological agreement. By anchoring adjudication on the scorer's judgment rather than on the LLM output's clinical merit, two reviewers could cover the full benchmark scope (69 cancer types) without requiring subspecialty expertise in every cancer. This aligns with the competency profile of medical oncology training, which covers the framework-level judgments (staging $\rightarrow$ pathway selection $\rightarrow$ information-gap recognition) common to all solid and hematologic malignancy management; such framework-level judgments remain feasible even outside a reviewer's subspecialty area.

\paragraph{Adjudication rubric.} Each item received one of seven verdict labels: \texttt{agree\_correct} / \texttt{agree\_partial} / \texttt{agree\_incorrect} (the scorer's verdict is clinically appropriate); \texttt{disagree\_should\_be\_correct} / \texttt{disagree\_should\_be\_partial} / \texttt{disagree\_should\_be\_incorrect} (the scorer's verdict is incorrect, with the clinician specifying the correct verdict); or \texttt{unsure} (insufficient clinical context to adjudicate). Free-text comments were optional. The full rubric and worked examples are provided in Supplementary~S1.

\paragraph{Reliability analysis.} Inter-rater agreement on the 25-item overlap set was quantified using Cohen's weighted $\kappa$~\cite{cohen1960kappa}, with verdicts treated as ordinal (\texttt{incorrect} $<$ \texttt{partial} $<$ \texttt{correct}) and linear weighting applied to off-diagonal disagreements. Scorer--clinician agreement was computed independently for each reviewer over their respective 125-item assignment (excluding \texttt{unsure} responses), again using Cohen's weighted $\kappa$. Ninety-five-percent confidence intervals were obtained by bootstrap resampling ($n = 2{,}000$, seed = 42). Results, confusion matrices, and per-disagreement analysis are reported in Section~\ref{sec:validation} and Supplementary~S1. Unlike multiple-choice benchmarks that inherit validity from established exam authorities (e.g., MedQA derived from USMLE~\cite{jin2021medqa}), and unlike rubric-on-output benchmarks that evaluate model responses rather than items (e.g., MedPaLM, MedPaLM 2, and HealthBench~\cite{singhal2023medpalm, singhal2025medpalm2, arora2025healthbench}), ODBB items require new validity evidence because they introduce a novel structured-decision verdict space (correct / partial / incorrect plus 14 typed failure labels). We provide this evidence via inter-rater $\kappa$ on a stratified item sample, which is, to our knowledge, the first such measurement reported for a clinical-LLM benchmark.

\paragraph{Gold-answer audit (second-wave validation).} The first-wave adjudication described above evaluates the scorer's \emph{verdict} given a fixed gold standard. A separate question is whether the gold standard itself is clinically defensible on the zero-correct items that underpin the 42\% boundary. We therefore conducted a second-wave audit specifically targeting gold-answer validity. From the 845 items in the zero-correct set, we drew a stratified random sample of $n = 80$ items (seed = 42; full sampling design in Supplementary~S7), oversampling the strata that reviewers might most plausibly question---all nine universal-fail items, 20 in-guide handoff, 15 upstream routing, 15 missing information, 15 parallel option, 3 evidence resolution, and 3 CRC. Two board-certified oncologists, independent of the first-wave adjudication (i.e., different individuals) and blinded from each other's responses, each re-evaluated every sampled gold answer along two orthogonal dimensions: \emph{clinical correctness} (is the NCCN gold answer clinically correct?) and \emph{NCCN traceability} (can the gold answer be located in the corresponding NCCN guideline workspace at the target pathway node?). Reviewers were not shown any model output, to avoid contamination by LLM responses. Inter-rater agreement, gold-error rate, and traceability rate were computed with Wilson 95\% confidence intervals for proportions, and Cohen's weighted $\kappa$ for the ordinal correctness scale; the second-wave audit results are reported in Section~\ref{sec:validation} and Supplementary~S7.

\subsection{Statistical analysis}\label{sec:stats}

\paragraph{Quantifying the collective capability boundary.} To test whether the 42\% boundary reflects a structural property of the model class or simply the limitation of any single model, we measured three complementary quantities. First, for each of the 2{,}005 items, we counted the number of models receiving a ``correct'' verdict (the per-item correctness count $k \in \{0, \ldots, 9\}$); items with $k = 0$ define the collective failure set. Second, we computed greedy-optimal ensemble coverage: starting from an empty model set, we repeatedly added the model whose inclusion covered the largest number of previously uncovered items, recording cumulative coverage at each step. This procedure yields an upper bound on what any single-pass ensemble can achieve. Third, to test whether ensemble saturation is an artifact of model similarity rather than a genuine capability ceiling, we computed pairwise verdict-level concordance across all 36 model pairs; high concordance combined with low coverage growth would indicate redundancy, whereas low concordance combined with low coverage growth indicates a shared blind spot.

\paragraph{Defining the knowledge-to-action gap.} Several models, particularly the more cautious open-weight ones, produced reasoning text that referenced the correct next clinical step yet stopped short of committing to it as the final decision. To quantify this pattern, we identified items in which the deterministic scorer flagged that the model's reasoning contained the gold-standard next step but the verdict was not ``correct,'' and we report the per-model rate of such items as the knowledge-to-action gap.

\paragraph{Statistical inference.} Reported model scores and pairwise comparisons carry uncertainty that single point estimates obscure. We therefore computed 95\% confidence intervals for NCCN strict concordance via bootstrap resampling of items~\cite{efron1993bootstrap} (2{,}000 iterations, seed = 42). Pairwise model comparisons used paired permutation tests over per-item differences (10{,}000 iterations), which make no distributional assumptions and respect the matched-pairs structure of identical item sets. To control the family-wise error rate across the eight comparisons of adjacently ranked models, we applied a Bonferroni correction setting $\alpha = 0.0063$. Reported proportions (the 42.1\% zero-correct rate and the gold-answer audit proportions) used Wilson score 95\% confidence intervals~\cite{wilson1927probable} rather than the normal approximation, because several proportions of interest lie near 0 or 1 where the normal interval is known to under-cover. All statistical analyses were conducted in Python 3.11.

\section{Results}\label{sec:results}

\subsection{Benchmark architecture overview}\label{sec:benchmark_overview}

\begin{figure*}[!t]
\centering
\includegraphics[width=\textwidth]{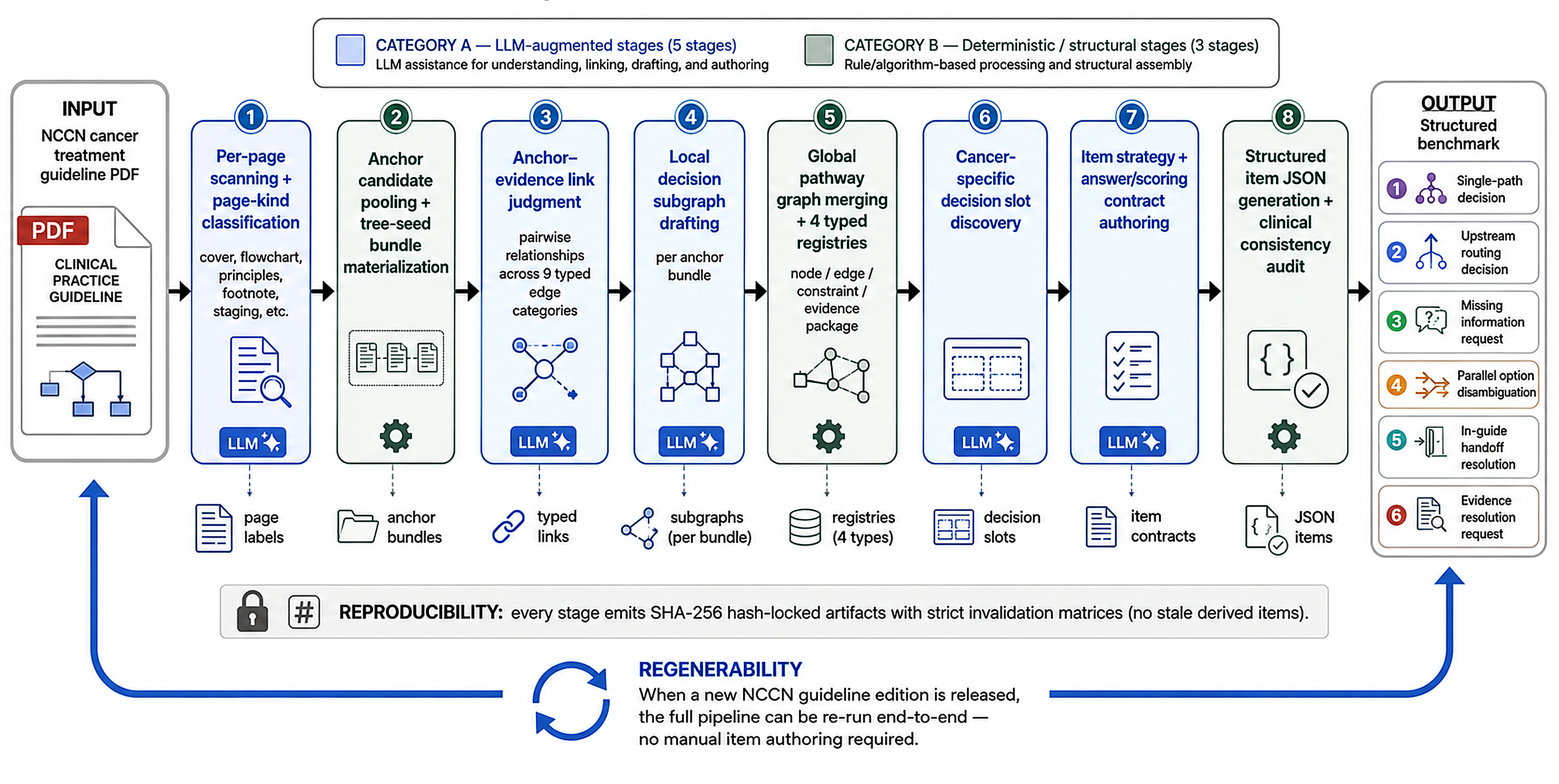}
\caption{Automated benchmark generation pipeline.}\label{fig:pipeline}
\begin{flushleft}
\small Note: The pipeline depicts production of the 1{,}586 NCCN structured items only; the 419 CRC case-based items are sourced separately via PubMed literature review (see ``CRC case sourcing'' paragraph in Methods Section~\ref{sec:benchmark}). Stages alternate between LLM-augmented semantic judgment (blue) and deterministic structural assembly (green); each emits SHA-256 hash-locked artifacts. NCCN guideline updates trigger full pipeline re-run.
\end{flushleft}
\end{figure*}

The Oncology Decision Boundary Benchmark comprises 2{,}005 scorable clinical decision items generated through an automated, fully reproducible pipeline (Figure~\ref{fig:pipeline}). The pipeline transforms a raw NCCN cancer treatment guideline PDF into a structured benchmark spanning six clinically motivated question types. Eight stages alternate between LLM-augmented semantic judgment (blue panels in Figure~\ref{fig:pipeline}; five stages comprising per-page classification, anchor--evidence link judgment, local subgraph drafting, cancer-specific slot discovery, and item strategy authoring) and deterministic structural assembly (green panels; three stages comprising anchor candidate pooling and tree-seed materialization, global graph merging followed by four typed registry construction, and structured item-JSON generation with clinical consistency audit). Every stage emits SHA-256 hash-locked artifacts with strict invalidation matrices, and the full pipeline can be re-run end-to-end to regenerate a fresh benchmark whenever NCCN releases a new guideline edition---no manual item authoring is required. The output is a structured benchmark of 1{,}586 NCCN guideline items plus 419 colorectal cancer case-based items, all schema-validated and provenance-traceable to specific NCCN pathway nodes or published case reports. A stratified random clinician audit on the hardest item subset confirms that 95\% of audited NCCN structured items are directly traceable to the cited NCCN pathway node (full audit reported in Section~\ref{sec:validation}, ``NCCN structured-item generation robustness''; Supplementary~S7), evidencing the robustness of the LLM-augmented item generation pipeline.

\subsection{Overall performance of nine frontier LLMs on ODBB: headline ranking, statistical tier separation, and cost-performance landscape}\label{sec:ranking}

Under the NCCN strict concordance metric (proportion of NCCN items receiving a \texttt{correct} verdict from the deterministic scorer; see Methods Section~\ref{sec:scorer} for the operational threshold), Claude Sonnet 4.6 ranked first (0.372), followed by GLM-5.1 (0.325), GLM-5 (0.322), Qwen 3.6-Plus (0.318), Minimax M2.7 (0.315), GPT-5.5 (0.302), DeepSeek V4 Pro (0.277), Gemini 2.5 Pro (0.230), and Gemini 3.1 Pro Preview (0.158) (Table~\ref{tab:ranking}). If the analysis stopped here, the conclusion would be straightforward: Claude Sonnet 4.6 is the best-performing model for clinical decision support. As we show below, this conclusion is misleading in multiple dimensions.

\begin{table}[ht]
\centering
\caption{Overall model performance on ODBB.}\label{tab:ranking}
\begin{tabular}{@{}rlcc@{}}
\toprule
Rank & Model & NCCN strict [95\% CI] & CRC strict [95\% CI] \\
\midrule
1 & Claude Sonnet 4.6 & 0.372 [0.349, 0.395] & 0.280 [0.238, 0.324] \\
2 & GLM-5.1 & 0.325 [0.303, 0.346] & 0.338 [0.295, 0.386] \\
3 & GLM-5 & 0.322 [0.298, 0.345] & 0.207 [0.172, 0.249] \\
4 & Qwen 3.6-Plus & 0.318 [0.298, 0.342] & 0.273 [0.232, 0.317] \\
5 & Minimax M2.7 & 0.315 [0.293, 0.337] & 0.234 [0.196, 0.277] \\
6 & GPT-5.5 & 0.302 [0.281, 0.327] & 0.250 [0.211, 0.294] \\
7 & DeepSeek V4 Pro & 0.277 [0.256, 0.301] & 0.206 [0.169, 0.247] \\
8 & Gemini 2.5 Pro & 0.230 [0.211, 0.252] & 0.198 [0.163, 0.239] \\
9 & Gemini 3.1 Pro Preview & 0.158 [0.141, 0.176] & 0.197 [0.163, 0.239] \\
\bottomrule
\end{tabular}
\begin{flushleft}
\small Note: NCCN strict concordance reported with bootstrap 95\% CI ($n = 2{,}000$ iterations); CRC strict concordance reported with Wilson 95\% CI on $n = 419$ CRC items.
\end{flushleft}
\end{table}

Permutation testing (10{,}000 iterations) revealed five statistically distinct performance tiers. Claude Sonnet 4.6 occupied a tier alone, significantly outperforming the second-ranked GLM-5.1 ($p = 0.005$). Models ranked 2nd through 6th (GLM-5.1, GLM-5, Qwen 3.6-Plus, Minimax M2.7, and GPT-5.5) formed an indistinguishable plateau: no pairwise comparison within this group reached significance (all $p > 0.05$; e.g., GLM-5.1 vs.\ GPT-5.5, $p = 0.13$ uncorrected and descriptive, not a Bonferroni-protected adjacency comparison). Below the plateau, three further tiers each contained a single model: DeepSeek V4 Pro (Tier~3), Gemini 2.5 Pro (Tier~4), and Gemini 3.1 Pro Preview (Tier~5), each separated from its immediate neighbor by Bonferroni-significant gaps (DeepSeek V4 Pro vs.\ Gemini 2.5 Pro $p < 0.001$; Gemini 2.5 Pro vs.\ Gemini 3.1 Pro Preview $p < 0.0001$). The GLM-5 to GLM-5.1 upgrade did not produce a statistically significant improvement ($p = 0.61$, the rank 3--4 adjacency comparison, Bonferroni-protected at $\alpha = 0.0063$). This tier structure undermines ranking-based model selection: choosing between the five plateau models on the basis of headline NCCN scores is not statistically justified. The practical implication is that if a clinical institution must choose among the five plateau models (GLM-5.1, GLM-5, Qwen 3.6-Plus, Minimax M2.7, GPT-5.5), headline accuracy provides no statistical basis---deployment decisions in this tier must instead be driven by other criteria such as cost, latency, decision-style preference, or local data-privacy constraints. Whether these five plateau models are substantively different or share common failure patterns is examined in the next subsection.

\begin{figure*}[!htbp]
\centering
\includegraphics[width=\textwidth]{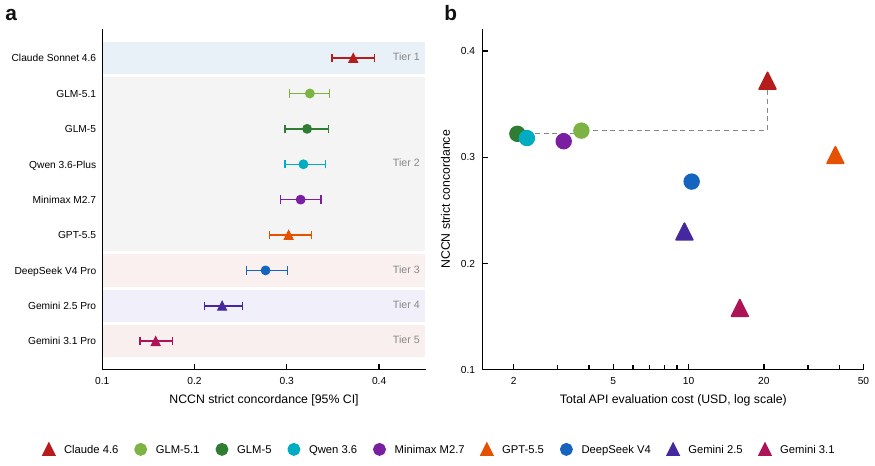}
\caption{Performance landscape of nine frontier LLMs on ODBB.}\label{fig:landscape}
\begin{flushleft}
\small Note: \textbf{a}, Forest plot of NCCN strict concordance with bootstrap 95\% confidence intervals (2{,}000 iterations). Permutation testing (10{,}000 iterations) identified five tiers: Tier~1 (Claude Sonnet 4.6, $p = 0.005$ vs.\ rank 2), Tier~2 (five-model plateau from GLM-5.1 to GPT-5.5; all pairwise $p > 0.05$), Tier~3 (DeepSeek V4 Pro), Tier~4 (Gemini 2.5 Pro), and Tier~5 (Gemini 3.1 Pro Preview). \textbf{b}, Cost--performance Pareto frontier (Section~\ref{sec:pareto}); $x$-axis = total API evaluation cost (USD, log scale), $y$-axis = NCCN strict concordance. The dashed step line connects the three Pareto-optimal models (GLM-5, GLM-5.1, Claude Sonnet 4.6); two of the three frontier models are open-weight (circles), and the remaining six models are all dominated at every price point. Triangles = closed-source; circles = open-weight.
\end{flushleft}
\end{figure*}

\subsection{The 42\% wall: a collective capability boundary}\label{sec:wall}

\begin{figure*}[!htbp]
\centering
\includegraphics[width=\textwidth]{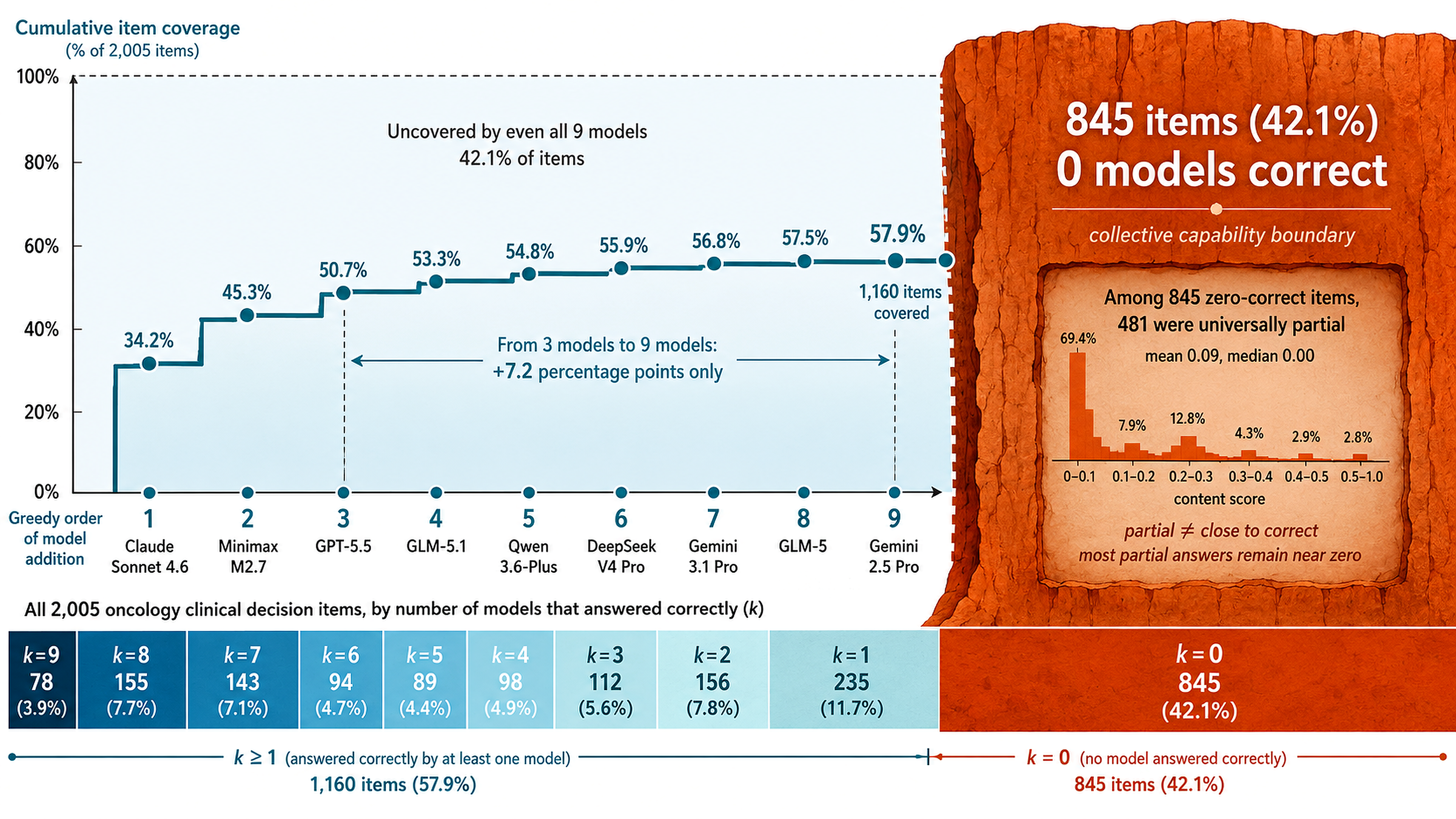}
\caption{The 42\% collective capability boundary.}\label{fig:wall}
\begin{flushleft}
\small Note: The bottom stacked bar shows the distribution of per-item correctness count ($k$) across 2{,}005 items: 845 items (42.1\%) at $k = 0$ (red block) versus 1{,}160 items (57.9\%) at $k \geq 1$. The upper-right inset within the highlighted ``wall'' panel shows the content score distribution for model--item pairs on $k = 0$ items ($n = 7{,}605$ pairs); 69.4\% of partial scores fall below 0.1 (mean 0.09, median 0.00), indicating that ``partial'' credit reflects marginal relevance rather than near-correct reasoning. The upper-left step curve shows greedy-optimal ensemble coverage as models are added one at a time; after three models, each additional model contributes less than 3 percentage points, yielding a nine-model ceiling of 57.9\%.
\end{flushleft}
\end{figure*}

Across all 2{,}005 items, 845 (42.1\%) were answered correctly by none of the nine models (the bottom per-item correctness distribution in Fig.~\ref{fig:wall}); this collective failure rate divided unevenly across the two tracks, with 35.7\% of NCCN guideline items (567/1{,}586) and 66.4\% of CRC case items (278/419) unsolved by any model. At the opposite extreme, only 78 items (3.9\%) were answered correctly by all nine models. The distribution of per-item correctness counts was strikingly bimodal, with 42.1\% of items at $k = 0$ and a secondary concentration at $k = 7$--$8$ (14.8\% combined), suggesting that clinical decision items tend to be either collectively solvable or collectively intractable.

Among the 845 zero-correct items, 481 (56.9\%) received a ``partial'' verdict from all nine models. Critically, the quality of these partial responses was very low: mean content score was 0.09 (median 0.00) on a 0--1 scale, with 69.4\% of model--item scores falling below 0.1 (the upper-right content-score histogram in Fig.~\ref{fig:wall}). ``Partial'' credit in these cases reflected the model touching on peripherally relevant information rather than approaching the correct clinical reasoning path.

To assess whether model diversity could breach this boundary, we computed greedy-optimal ensemble coverage (the upper-left step curve in Fig.~\ref{fig:wall}). The best single model (Claude Sonnet 4.6) covered 34.2\% of items. Adding Minimax M2.7 raised coverage to 45.3\% (+11.2 percentage points), and adding GPT-5.5 reached 50.7\% (+5.3 pp). From the fourth model onward, marginal gains declined sharply: models 4 through 9 collectively contributed only 7.2 additional percentage points, yielding a nine-model ceiling of 57.9\%. This rapid saturation indicates that the remaining 42.1\% of items represent a qualitatively different category of difficulty---one that cannot be addressed by adding more models of the current generation~\cite{bommasani2023helm}. The 57.9\% ceiling has a direct deployment implication: ensemble-based clinical reasoning strategies, which prior work has proposed as a path to higher reliability, face an inherent ceiling at this generation of models. Marginal gains from adding the 4th--9th model average only 1.2 percentage points each, below the threshold that would justify the additional API cost for most institutional deployments. Having shown that 42\% of items defeat the entire model class, we now dissect what makes them collectively intractable.

\subsection{Anatomy of the wall: cross-pathway navigation as the core blind spot}\label{sec:anatomy}

To understand what makes the zero-correct items collectively intractable, we analyzed their composition by track and question type (Table~\ref{tab:zerocorr}). Two strata stood out: in-guide handoff resolution (77.2\% zero-correct rate, 105/136 items) and upstream routing decision (64.5\%, 71/110 items)---both reflecting cross-pathway navigation. CRC case-based therapy recommendation also showed a high zero-correct rate (66.3\%, 278/419), reflecting the difficulty of multi-line systemic therapy selection from free-text clinical scenarios. By contrast, parallel option disambiguation---a within-pathway reasoning task---had only 21.7\% zero-correct.

\begin{table}[ht]
\centering
\caption{Distribution of 845 zero-correct items across question types.}\label{tab:zerocorr}
\begin{tabular}{@{}lrrr@{}}
\toprule
Question type & Zero-correct & Total & \% \\
\midrule
In-guide handoff resolution & 105 & 136 & 77.2 \\
Single path decision & 2 & 3 & 66.7 \\
Upstream routing decision & 71 & 110 & 64.5 \\
Missing information request & 165 & 348 & 47.4 \\
Evidence resolution request & 24 & 69 & 34.8 \\
Parallel option disambiguation & 200 & 920 & 21.7 \\
\midrule
NCCN subtotal & 567 & 1{,}586 & 35.7 \\
CRC therapy recommendation & 278 & 419 & 66.3 \\
\midrule
\textbf{Total} & \textbf{845} & \textbf{2{,}005} & \textbf{42.1} \\
\bottomrule
\end{tabular}
\begin{flushleft}
\small Note: NCCN structured items appear above the midrule; CRC items below.
\end{flushleft}
\end{table}

The pattern was most stark for universal failures. Nine NCCN items were answered incorrectly (not merely partially) by all nine models; all nine fell into a cross-pathway navigation stratum (in-guide handoff or upstream routing), requiring the model to recognize that the patient's clinical scenario demanded routing to a different guideline pathway before proceeding (Table~\ref{tab:universal_fail}). Given the small subgroup size ($n = 9$), we treat this as illustrative rather than independently load-bearing; the primary evidence that cross-pathway navigation drives the boundary comes from the broader zero-correct distribution (Table~\ref{tab:zerocorr}: in-guide handoff 77.2\% zero-correct, upstream routing 64.5\%). This represents a systematic gap in what we term \emph{meta-decision competence}: the ability to determine which reasoning framework applies before reasoning within it. This mirrors a well-known taxonomy of diagnostic error~\cite{graber2005diagnostic}: premature closure and context errors arise precisely when clinicians fail to reconsider the applicable reasoning framework.

\begin{table}[ht]
\centering
\small
\caption{The nine NCCN items where all nine LLMs produced \texttt{incorrect} verdicts.}\label{tab:universal_fail}
\begin{tabular}{@{}lll@{}}
\toprule
Clinical domain & Question type & Failure family \\
\midrule
HIV-associated cancer & In-guide handoff & Ambiguous routing target \\
Cancer of unknown primary & In-guide handoff & Ambiguous routing target \\
Primary cutaneous lymphoma & In-guide handoff & Ambiguous routing target \\
Primary cutaneous lymphoma & In-guide handoff & Ambiguous routing target \\
Older adult oncology & In-guide handoff & Ambiguous routing target \\
Non-small cell lung cancer & In-guide handoff & Ambiguous routing target \\
Pediatric ALL & Upstream routing & Cross-guideline transition \\
Rectal cancer & Upstream routing & Cross-guideline transition \\
Rectal cancer & Upstream routing & Cross-guideline transition \\
\bottomrule
\end{tabular}
\begin{flushleft}
\small Note: All cases involve cross-pathway navigation (in-guide handoff or upstream routing).
\end{flushleft}
\end{table}

By question type, model performance on handoff and routing tasks was uniformly poor (Table~\ref{tab:qtype}). The best model on in-guide handoff resolution achieved only 14.0\% correct (GPT-5.5), while six of nine models scored below 3\%. For upstream routing decisions, the best model reached 19.1\% correct (GPT-5.5), with five models below 7\%. In contrast, for parallel option disambiguation---a within-pathway reasoning task---seven of nine models exceeded 40\% strict concordance. If models can solve within-pathway reasoning (POD at 40\%+) but fail at cross-pathway navigation (IGHR $<$ 15\% for most models), the bottleneck is not clinical knowledge itself but the meta-cognitive judgment of \emph{which} pathway to apply. This points to a qualitatively different capability gap than what training-data scaling addresses. Meta-decision competence is one form of decision difficulty; the next two subsections explore two others---decisiveness calibration and the knowledge-to-action transition.

\begin{table}[ht]
\centering
\caption{Strict concordance rate (\%) by question type and model.}\label{tab:qtype}
\begin{tabular}{@{}lccccc@{}}
\toprule
Model & POD & MIR & ERR & IGHR & URD \\
\midrule
Claude Sonnet 4.6 & \textbf{49.7} & \textbf{24.1} & 17.4 & 13.2 & 17.3 \\
GPT-5.5 & 40.0 & 12.9 & 36.2 & \textbf{14.0} & \textbf{19.1} \\
Gemini 2.5 Pro & 32.2 & 13.2 & 29.0 & 1.5 & 0.9 \\
Gemini 3.1 Pro Preview & 21.3 & 6.9 & 14.5 & 7.4 & 10.0 \\
Qwen 3.6-Plus & 45.9 & 16.7 & 31.9 & 0.0 & 2.7 \\
DeepSeek V4 Pro & 40.7 & 11.3 & 26.5 & 2.2 & 6.4 \\
GLM-5 & 45.9 & 17.0 & 37.7 & 1.5 & 1.8 \\
GLM-5.1 & \textbf{47.8} & 16.1 & 20.3 & 2.2 & 2.7 \\
Minimax M2.7 & 43.6 & 18.1 & \textbf{44.8} & 2.2 & 9.6 \\
\bottomrule
\end{tabular}
\begin{flushleft}
\small Note: Bold values indicate best per question type. POD = parallel option disambiguation; MIR = missing information request; ERR = evidence resolution request; IGHR = in-guide handoff resolution; URD = upstream routing decision.
\end{flushleft}
\end{table}

\begin{figure*}[!htbp]
\centering
\includegraphics[width=\textwidth]{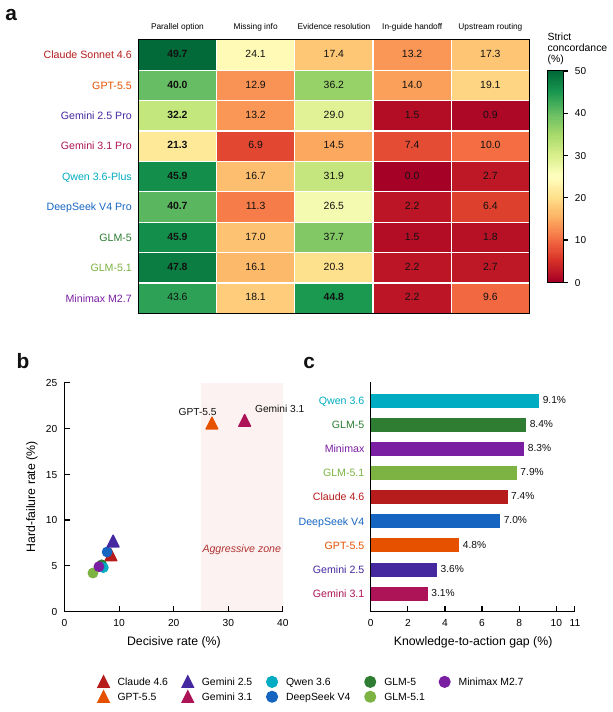}
\caption{Anatomy of LLM failure modes on ODBB.}\label{fig:anatomy}
\begin{flushleft}
\small Note: \textbf{a}, Question-type performance heatmap; strict concordance (\%) by model $\times$ question type on NCCN items. IGHR (in-guide handoff resolution) and URD (upstream routing decision) columns show uniformly low scores (dark shading), indicating cross-pathway navigation is a shared blind spot. Bold values indicate best performance per question type. \textbf{b}, Decisiveness--safety tradeoff (Section~\ref{sec:tradeoff}); $x$-axis = decisive rate (\%), $y$-axis = hard-failure rate (F1 unsafe overreach + F3 premature commitment, \%). GPT-5.5 and Gemini 3.1 Pro Preview occupy the aggressive zone (red shading), exhibiting three- to fivefold higher hard-failure rates than the seven conservative models. \textbf{c}, Knowledge-to-action gap by model (Section~\ref{sec:kta}); proportion of items where the model's reasoning mentioned the correct next clinical step but its final decision did not reflect that knowledge. Conservative open-weight models show larger gaps (between 7\% and 9\%); the aggressive Gemini 3.1 Pro Preview shows a smaller gap (3.1\%) only because it commits more readily---often to incorrect actions.
\end{flushleft}
\end{figure*}

\subsection{The cost of climbing the wall: decisiveness versus safety}\label{sec:tradeoff}

Two models---GPT-5.5 and Gemini 3.1 Pro Preview---adopted markedly more aggressive decisiveness strategies than the remaining seven (Fig.~\ref{fig:anatomy}B). Their combined decisive rates (proportion of items where the model chose ``proceed'' or ``routing'' rather than requesting more information) were 27.0\% and 33.0\%, respectively, compared with 5.2\% to 9.0\% for the conservative majority.

This aggressiveness came at a steep safety cost. GPT-5.5 exhibited a hard-failure rate (F1 unsafe overreach + F3 premature commitment) of 20.6\% (Wilson 95\% CI 18.7\%--22.7\%, $n=1{,}586$ NCCN items), and Gemini 3.1 Pro Preview 20.9\% (Wilson 95\% CI 18.9\%--22.9\%), compared with 4.2--7.7\% for the seven conservative models---an approximately three- to fivefold difference.

The Gemini version comparison provided a particularly instructive case study. Gemini 3.1 Pro Preview scored 0.158 on NCCN strict concordance, a 31\% decline from Gemini 2.5 Pro's 0.230, despite being marketed as an improved model with enhanced agentic capabilities. Disaggregating by failure label revealed the mechanism: F1 (unsafe overreach) counts rose from 56 to 158 ($2.8\times$), F3 (premature commitment) from 48 to 173 ($3.6\times$), and the model's use of ``stop---need evidence'' decisions collapsed from 155 to 4 ($-97\%$). The model became more decisive but less safe: precisely the wrong tradeoff for clinical applications where approximately 84\% of NCCN items (1{,}337 of 1{,}586---across parallel-option, missing-information, and evidence-resolution question types) require the model to request additional information before acting. This version-to-version regression echoes broader concerns about LLM behavioral drift across updates~\cite{chen2023chatgptdrift, ma2023regression}. The deployment implication is that ``more decisive'' is not a free upgrade: two of the most recent and heavily-tuned models in our sample are also the two least safe by clinical-action standards. This argues against using vendor-reported ``agentic capability'' as a proxy for clinical readiness. Aggressive decisiveness commits to action without sufficient evidence; the opposite pathology---committing to \emph{no} action despite having sufficient evidence---has equally important clinical implications, which we examine next.

\subsection{The knowledge-to-action gap}\label{sec:kta}

In a fraction ranging from 3.1\% to 9.1\% of items across models, the deterministic scorer identified that the model's reasoning text mentioned the correct next clinical step yet the model's final decision did not reflect this knowledge. We term this the ``knowledge-to-action gap'': the model possesses the relevant clinical knowledge but fails to translate it into a committed clinical decision. This phenomenon may be partially attributable to LLM overconfidence calibration failures, whereby models express high certainty in their reasoning yet decline to act on it~\cite{savage2025llmcalibration, xiong2024llmconfidence}.

The gap was most pronounced in the most cautious models: Qwen 3.6-Plus (9.1\%), GLM-5 (8.4\%), and Minimax M2.7 (8.3\%). Two clinical scenario types most commonly triggered the gap. First, when the guideline permitted two or three parallel acceptable regimens, the model would enumerate the correct option in its reasoning but then decline to commit, instead requesting ``additional information''---a pattern consistent with the model deferring judgment when no single option carries decisive evidence. Second, when therapy selection required integrating multiple biomarkers, the model would identify the key discriminator in its reasoning (e.g., RAS/BRAF status determining EGFR-targeted therapy eligibility) but then present multiple regimens in parallel rather than selecting one.

Aggressive models showed smaller gaps (Gemini 3.1 Pro Preview: 3.1\%) \emph{not} because they were better at translating knowledge to action, but because they committed to \emph{some} action more readily---often the wrong one (cf.\ Section~\ref{sec:tradeoff}). These results indicate that, on clinical LLM tasks, ``action rate'' and ``accuracy'' are not synonyms; they may even be inversely correlated.

The finding has a concrete implication for model training. Current RLHF objectives typically reward task completion or user satisfaction, indirectly encouraging models to commit to a final decision. But the reward structure in clinical practice is asymmetric: the cost of a wrong commitment far exceeds the cost of ``I recommend further consultation.'' Incorporating ``when not to commit'' as an explicit capability dimension during fine-tuning---for instance, expected-cost-aware reward shaping for medical settings---is a feasible path to closing the knowledge-to-action gap. The 3.1\%--9.1\% knowledge-to-action gap quantifies an upper-bound improvement reachable through reward-shaping interventions alone, \emph{without requiring any new clinical knowledge}. Claude Sonnet 4.6 demonstrated yet a third behavioral pattern---the highest accuracy combined with the highest contraindicated-recommendation rate---warranting closer examination in the next section.

\subsection{Within the wall: rank reversals and irreplaceability}\label{sec:reversals}

The two-track design tests a question that NCCN performance alone cannot answer: whether a model retains the expert-clinician-like capacity to produce contextually-correct decisions that deviate from consensus guidance. Mature physicians routinely make such deviations when patient context warrants, and published case reports systematically over-represent these cases precisely because the treating physician's expertise added value beyond the guideline. A model that scores well on CRC despite weaker NCCN performance is, by construction, reproducing this kind of context-driven deviation rather than mainstream consensus. Table~\ref{tab:reversals} provides direct evidence that this capacity is dissociable from---and can survive---weaker NCCN alignment: GPT-5.5 (NCCN 6th $\to$ CRC 4th), Gemini 2.5 Pro (8th $\to$ 6th), and Gemini 3.1 Pro Preview (9th $\to$ 7th) all gained rank from NCCN to CRC despite mid-to-low NCCN positions. The reverse direction is exemplified by GLM-5 (NCCN 3rd, CRC 9th, a six-position drop that persists at strict drug-matching granularity, strict therapy F1@3 of 0.207 vs.\ GLM-5.1's 0.338): top-tier consensus mastery did not transfer to case-level deviation reasoning. GLM-5.1 simultaneously held 2nd on NCCN and 1st on CRC, demonstrating that within-vendor version differences can outweigh cross-vendor differences for specific task tracks. The deployment consequence is symmetric: NCCN benchmark performance is neither necessary nor sufficient as a proxy for a model's case-level deviation-reasoning capacity, and single-benchmark procurement decisions are vulnerable in both directions.

Individual models also demonstrated unique solving capabilities. Minimax M2.7 uniquely solved 57 items that no other model answered correctly, followed by GPT-5.5 with 52 unique solves (Table~\ref{tab:reversals}). Neither model ranked in the top three overall, yet both were the least replaceable members of any ensemble. Each model also showed a distinct question-type specialty: Claude Sonnet 4.6 and the open-weight cluster excelled at parallel option disambiguation, while Minimax M2.7 uniquely topped the evidence resolution request type (44.8\% strict concordance vs.\ 37.7\% for the next-best model). This divergence between aggregate ranking and marginal contribution underscores that clinical model selection based solely on headline accuracy metrics systematically undervalues models with distinctive reasoning profiles.

\begin{table}[ht]
\centering
\caption{Per-model NCCN rank, CRC rank, rank change ($\Delta$), unique-solve count, and top question-type specialty.}\label{tab:reversals}
\begin{tabular}{@{}lcccrl@{}}
\toprule
Model & NCCN rank & CRC rank & $\Delta$ & Unique solves & Top specialty \\
\midrule
Claude Sonnet 4.6 & 1 & 2 & $\downarrow$1 & 40 & POD (49.7) \\
GLM-5.1 & 2 & 1 & $\uparrow$1 & 19 & POD (47.8) \\
GLM-5 & 3 & 9 & $\boldsymbol{\downarrow 6}$ & 13 & POD (45.9) \\
Qwen 3.6-Plus & 4 & 3 & $\uparrow$1 & 20 & POD (45.9) \\
Minimax M2.7 & 5 & 5 & --- & \textbf{57} & ERR (44.8) \\
GPT-5.5 & 6 & 4 & $\uparrow$2 & \textbf{52} & POD (40.0) \\
DeepSeek V4 Pro & 7 & 8 & $\downarrow$1 & 14 & POD (40.7) \\
Gemini 2.5 Pro & 8 & 6 & $\uparrow$2 & 7 & POD (32.2) \\
Gemini 3.1 Pro Preview & 9 & 7 & $\uparrow$2 & 13 & POD (21.3) \\
\bottomrule
\end{tabular}
\begin{flushleft}
\small Note: NCCN rank by NCCN strict concordance; CRC rank by class-level therapy F1@3. POD = parallel option disambiguation; ERR = evidence resolution request.
\end{flushleft}
\end{table}

The top-ranked model, Claude Sonnet 4.6, exhibited its own safety paradox. While achieving the highest overall score (0.372), it produced the highest CRC contraindicated-recommendation rate (6.0\%---i.e., in 6\% of CRC items it recommended at least one drug explicitly contraindicated by the gold standard), and the lowest rate of acknowledging ``data conflicts'' in its reasoning text (2 instances vs.\ 24 for Qwen 3.6-Plus). Simultaneously, Claude showed the lowest contradictory-output rate (1.4\% vs.\ 10.6\% for GPT-5.5).

Read together, these numbers suggest a mechanistic hypothesis: Claude's training appears to have optimized for internal output coherence and commitment confidence, but at the cost of reduced sensitivity to data contradictions. When clinical scenarios contain latent conflicts (e.g., guideline recommendation versus patient preference, new evidence versus standard regimen), Claude tends to produce an internally coherent answer rather than pausing to flag the contradiction; this is a feature on USMLE-style single-answer tests (the answer reads confident) but in clinical practice it amounts to having the model silently absorb gray-zone judgment risk.

This observation reflects a broader principle: high accuracy and high safety are not interchangeable constructs~\cite{asgari2025llmsafety, chen2023chatgptcancer}. In fact, Claude's case suggests they can trade off against each other---making the model more confident about its own outputs can make it less alert to contradictions in the input. Clinical-deployment procurement processes therefore need to evaluate ``sensitivity to contradiction'' and ``commitment strength in low-uncertainty settings'' as two independent dimensions, rather than collapsing both into a single headline-accuracy number. Beyond clinical content, the deployment decision also turns on cost and access; the next section analyzes the cost-performance frontier.

\subsection{Cost-effectiveness and the open-weight Pareto frontier}\label{sec:pareto}

Three of the nine evaluated models lay on the cost-performance Pareto frontier (per-model total API evaluation costs computed from actual token usage during the May 2026 run; see Methods Section~\ref{sec:models}). Two of the three frontier models were open-weight: GLM-5 (\$2.07, 0.322) and GLM-5.1 (\$3.73, 0.325). Claude Sonnet 4.6 was the sole closed-source model on the frontier (\$20.68, 0.372)---achieving a 14.5\% score advantage at 5.5$\times$ the cost of GLM-5.1, or a 15.5\% advantage at 10.0$\times$ the cost of GLM-5. The remaining six models---GPT-5.5 (\$38.60), Gemini 3.1 Pro Preview (\$16.03), DeepSeek V4 Pro (\$10.29), Gemini 2.5 Pro (\$9.63), Minimax M2.7 (\$3.17), and Qwen 3.6-Plus (\$2.26)---were all dominated: at every price point, GLM-5, GLM-5.1, or Claude Sonnet 4.6 offered equal or better performance.

Four of the five open-weight models also exhibited behavioral clustering, with pairwise verdict-level concordance exceeding 0.77 for GLM-5.1/Qwen 3.6-Plus (0.801), GLM-5.1/GLM-5 (0.800), and GLM-5/Qwen 3.6-Plus (0.778). This convergence suggests that the open-weight training ecosystem may have developed a shared orientation toward conservative decision-making that, while limiting decisiveness, confers lower hard-failure rates (4.2--7.7\% vs.\ 20.6--20.9\% for the aggressive closed-source models). This convergence is not merely a cost story but constitutes a \emph{behavioral fingerprint} distinguishing the open-weight cluster from the aggressive closed-source cluster. For clinical-deployment audit, this means open-weight models offer not only auditable weights and local-deployment options but also a more predictable safety profile---a triplet of advantages that is difficult to obtain from closed-source frontier models at any price point. All preceding findings nevertheless hinge on the deterministic scorer's verdicts being clinically defensible; the next section provides direct evidence on this point.

\subsection{Clinician validation}\label{sec:validation}

Two board-certified oncologists independently reviewed 225 stratified-sample items (125 per oncologist, 25 overlapping). Five judgments were marked ``unsure'' or left blank and were excluded from analysis, yielding 245 evaluable scorer--clinician pairs.

Overall, the oncologists agreed with the scorer's verdict on 228 of 245 evaluable pairs (93.1\%; pair-level agreement). Agreement was high across all verdict categories: 95.3\% for ``correct'' items (Doctor~A) and 85.0\% (Doctor~B), 97.3\% and 91.0\% for ``partial'' items, and 100\% for ``incorrect'' items from both oncologists. Scorer--clinician Cohen's weighted $\kappa$ (linear weights, $w_{ij} = 1 - |i-j|/(k-1)$) was 0.939 (95\% bootstrap CI [0.870, 0.985]) for Doctor~A and 0.790 (95\% bootstrap CI [0.671, 0.892]) for Doctor~B, both indicating substantial-to-almost-perfect agreement~\cite{cohen1960kappa}.

On the 25 overlapping items used for inter-rater reliability, the two oncologists agreed on 22 of 25 items (88.0\%), yielding an inter-rater weighted $\kappa$ of 0.757 (substantial agreement). All three disagreements involved items where the scorer assigned ``partial'' and Doctor~A concurred but Doctor~B upgraded to ``correct''---a pattern consistent with Doctor~B applying a slightly more lenient standard for partial-credit items rather than any systematic scorer error.

These results establish that the deterministic scorer produces clinically defensible verdicts: its judgments agree with independent oncologist review at rates comparable to inter-rater agreement between the oncologists themselves. In particular, the scorer's ``incorrect'' verdicts---which directly underpin the 42\% capability boundary---received unanimous clinician endorsement in the adjudication sample (9/9), arguing that the 42\% boundary reflects genuine LLM failure rather than scorer false positives.

\paragraph{NCCN structured-item generation robustness and gold-answer validity on the zero-correct subset.} A separate concern is whether the LLM-augmented item generation pipeline produces clinically defensible structured items---that is, whether the items themselves, and their gold answers, on the zero-correct subset can withstand independent clinician scrutiny, or whether the 42\% wall could in principle reflect generation artifacts or bad gold rather than genuine LLM failure. We addressed this with a second-wave audit (Section~\ref{sec:clinician}, ``Gold-answer audit''; full design in Supplementary~S7) in which two board-certified oncologists independently re-evaluated a stratified random sample of $n = 80$ items drawn from the 845 zero-correct items, scoring each gold answer along two orthogonal dimensions: \emph{clinical correctness} (is the gold answer clinically correct?) and \emph{NCCN traceability} (can the gold answer be located in the corresponding NCCN guideline workspace at the target pathway node?). Both reviewers labeled the gold answer \texttt{traceable} for 76 of 80 items (95.0\%, Wilson 95\% CI 87.8\%--98.0\%), and both labeled the gold answer \texttt{wrong} for only 2 of 80 items (2.5\%, Wilson 95\% CI 0.7\%--8.7\%); under the most permissive definition (either reviewer flagged \texttt{wrong}), the bound is 3 of 80 (3.8\%, Wilson 95\% CI 1.3\%--10.5\%). Critically, none of the nine universal-fail items underpinning the hardest subset of the 42\% wall received a \texttt{wrong} flag from either reviewer (0/9); they were unanimously labeled \texttt{correct} or \texttt{arguable}. These results bound the gold-error rate on the zero-correct subset well below the threshold at which the 42\% capability boundary could be re-explained as a gold-quality artifact, supporting the interpretation that the boundary reflects genuine LLM failure rather than LLM-introduced item-generation errors or weak gold standards.

\section{Related work}\label{sec:related}

Clinical LLM evaluation has matured along three lines~\cite{chang2024survey, bommasani2023helm, thirunavukarasu2023llmmedicine, clusmann2023llmclinical}, each with distinct validity foundations and evaluation modalities. We position ODBB against these lines and identify the specific gap it fills.

\paragraph{Knowledge-recall benchmarks.} MedQA~\cite{jin2021medqa}, PubMedQA~\cite{jin2019pubmedqa}, the broader USMLE-derived benchmark family~\cite{nori2023gpt4, kung2023chatgpt, hendrycks2021mmlu, lievin2024medqa}, and large-scale aggregations such as MedBench~\cite{mcduff2023medbench} test factual recall via single-answer multiple choice. These benchmarks inherit validity from established medical-exam authorities (USMLE, board certifications), eliminating the need for new item-level reliability studies but constraining the evaluation signal to fact retrieval rather than to multi-step decision-path navigation~\cite{topol2019highperformance, sutton2020cdss}. Prompt-engineering studies such as Medprompt~\cite{nori2023gpt4medprompt} and recent foundation-model entries including Med-Gemini~\cite{saab2024medgemini}, Meditron~\cite{chen2023meditron}, MedAlpaca~\cite{han2023medalpaca}, and OpenBioLLM~\cite{pal2024openbiollm} have been evaluated primarily against this same benchmark substrate. MultiMedQA~\cite{singhal2023medpalm} extended the lineage by aggregating MedQA with open-ended consumer-health questions and providing the first large-scale (14-clinician, 7-axis) evaluation of LLM responses---though that evaluation targeted model outputs rather than benchmark items themselves.

\paragraph{Rubric-on-output benchmarks.} MedPaLM and MedPaLM 2~\cite{singhal2023medpalm, singhal2025medpalm2} pioneered the rubric-based evaluation paradigm for open-ended medical responses, in which clinicians score model answers along axes such as scientific consensus, potential harm, bias, and helpfulness. HealthBench~\cite{arora2025healthbench} subsequently extended this paradigm to a substantially larger physician-graded ground-truth set across multiple medical specialties, while retaining the response-level rubric scoring methodology. Similar rubric-based methodologies have been adopted in oncology-specific evaluations~\cite{li2026cancerllm, hao2025oncollm, carl2024llmoncology, chen2023chatgptcancer} and in broader holistic-evaluation frameworks~\cite{bommasani2023helm, srivastava2023beyond, agrawal2025evaluation}. Hallucination-focused benchmarks~\cite{pal2023medhalt, ji2023hallucination} and safety-oriented evaluations~\cite{asgari2025llmsafety} use related output-grading methodologies. These methods are well-suited to assessing open-ended response quality but inherit per-evaluation costs from human or LLM-judge involvement~\cite{zheng2023judging}, and they report inter-rater agreement primarily on \emph{model outputs}, not on benchmark \emph{items} themselves.

\paragraph{Decision-oriented and domain-specific benchmarks.} Diagnostic-pathway and oncology-focused benchmarks~\cite{hao2025oncollm, carl2024llmoncology, sorin2023gptbreast, griewing2024breastllm}, clinical-decision-support evaluation frameworks~\cite{sutton2020cdss, rajpurkar2022aihealth, gaube2021humanai}, multi-agent and ensemble approaches~\cite{tang2023medagents, kim2024mdagents, bousquet2025multiagent}, retrieval-augmented architectures~\cite{lewis2020rag, zakka2024almanac, lewis2025ragclinical}, and calibration-focused evaluations~\cite{savage2025llmcalibration, xiong2024llmconfidence} all target structured clinical tasks closer to real practice. Scoring is, however, typically binary (\texttt{correct} versus \texttt{incorrect} against a single gold answer), which collapses the partial-credit failures, decisiveness-safety tradeoffs, and knowledge-to-action gaps that turn out to be central to clinical-deployment risk (Section~\ref{sec:results}). Cross-vendor regression and behavioral-drift studies~\cite{chen2023chatgptdrift, ma2023regression} further motivate a reproducibility requirement that single-correct-answer scoring cannot itself guarantee. Cross-system clinical databases~\cite{wang2025csedb} highlight the importance of grounding benchmarks in authoritative guideline sources such as NCCN~\cite{nccn2025guidelines}.

\paragraph{Test-time reasoning evaluations.} Recent evaluations of OpenAI's o-series reasoning models~\cite{nori2024o1medical, openai2024o1systemcard} demonstrate that test-time reasoning can substantially improve MedQA-style benchmark performance compared with non-reasoning chat models from the same family. These evaluations remain anchored to single-answer multiple-choice scoring inherited from MedQA~\cite{jin2021medqa}, however, and have not been systematically evaluated against decision-path navigation tasks that require the model to commit to one of several internally permissible pathways before reasoning within any. ODBB extends the comparison surface to this latter, decision-path setting.

\paragraph{ODBB's positioning and the gap it fills.} ODBB extends this landscape in three orthogonal directions. \textbf{First}, it introduces a structured-decision verdict space (\texttt{correct} / \texttt{partial} / \texttt{incorrect} plus 14 typed failure labels) sufficient to characterize the full failure-mode distribution that single-correct-answer benchmarks collapse into a binary label. \textbf{Second}, it eliminates LLM-as-judge inference entirely from the scoring pipeline~\cite{zheng2023judging, bommasani2023helm, agrawal2025evaluation}, producing bit-for-bit reproducible verdicts that support longitudinal model comparison without judge drift~\cite{chen2023chatgptdrift, ma2023regression}. \textbf{Third}, it performs clinician adjudication on \emph{scorer judgments} rather than on model outputs, yielding---to our knowledge---the first item-level inter-rater $\kappa$~\cite{cohen1960kappa} reported for a clinical-LLM benchmark. The combination of determinism and clinician validation also enables downstream architectural applications (boundary-classifier training, expected-cost-aware reward shaping~\cite{ouyang2022training}) that no prior benchmark supports at the verdict-space granularity required.

\section{Discussion}\label{sec:discussion}

\subsection{From ranking to boundary}\label{sec:discussion-boundary}

The dominant paradigm in clinical LLM evaluation has been comparative ranking: testing models on a benchmark, sorting by score, and recommending the top performer for deployment. This paradigm carries an implicit assumption---that for any clinical task, some model in the current generation is good enough, and the evaluator's job is to find it~\cite{chang2024survey, nori2023gpt4medprompt}.

Our data are inconsistent with this assumption. The 42.1\% collective failure rate, the rapid saturation of ensemble coverage at 57.9\%, and the uniformly poor quality of partial responses on zero-correct items together indicate that a substantial fraction of clinical decision-making appears to lie beyond the current capability frontier of the entire model class. This pattern is better described as a boundary problem than as a ranking problem.

The reframing has a concrete architectural consequence. ``Boundary-aware'' clinical deployment has at least three identifiable components: (i) a \emph{boundary classifier} that runs alongside the primary clinical model and predicts, per item, whether the item is likely to fall within the model's capability envelope or beyond it; (ii) a \emph{specialist-routing} layer that, for items deemed within-envelope, selects the most appropriate model from a portfolio based on question type (e.g., parallel-option disambiguation versus upstream routing); and (iii) a \emph{structured triage protocol} that, for items predicted to fall beyond the envelope, escalates to a human clinician with the model's reasoning attached, so the clinician can review the model's partial work rather than restart from scratch~\cite{rajpurkar2022aihealth, gaube2021humanai}. None of these three components is a model improvement; all three are evaluation-and-architecture investments that the current sub-field roadmap has under-prioritized relative to model-scale and agentic-capability investments.

Our methodological contributions support exactly this reframing. The deterministic scorer with 14 typed failure labels provides the substrate on which a boundary classifier could be trained (the failure-label distribution is a natural feature space). The clinician adjudication (scorer--clinician weighted $\kappa = 0.790$--$0.939$) supports the claim that scorer verdicts are clinically defensible, which is a prerequisite for any boundary classifier built on top of those verdicts. The benchmark generation pipeline, by being end-to-end re-runnable, supports the longitudinal re-evaluation that boundary-aware deployment requires as both guidelines and models evolve.

\subsection{Three failure modes that constitute the boundary}\label{sec:discussion-failure-modes}

Having established that 42\% of items lie beyond the collective capability frontier, the next question is what \emph{kinds} of failure constitute the boundary. Three distinct failure modes emerged from the analysis, each with a distinct training- or architecture-level corrective direction.

\textbf{Failure mode 1: cross-pathway navigation as a meta-judgment gap.} All nine models exhibited disproportionately poor performance on in-guide handoff resolution and upstream routing decision items (Section~\ref{sec:anatomy}). This reflects a structural limit of current training paradigms: standard pretraining and RLHF~\cite{ouyang2022training} optimize for within-context reasoning---given a document or conversation, generate a coherent continuation. Cross-pathway navigation, by contrast, demands multi-pathway meta-reasoning: determining \emph{which} pathway applies before reasoning \emph{within} it. This parallels a well-documented pattern in human clinical cognition. Specialist physicians display fluent within-specialty reasoning but require deliberate, slower processes for cross-specialty consultation~\cite{elstein1978medical, lambe2016dual}; multi-disciplinary case conferences exist precisely because cross-domain routing is cognitively distinct from within-domain reasoning. Current LLM architectures lack an analogous mechanism for recognizing when the active reasoning context is insufficient. Resolving this likely requires either retrieval-augmented architectures~\cite{lewis2020rag, zakka2024almanac, lewis2025ragclinical} that explicitly route queries to the correct pathway before reasoning, structured prompting that decomposes routing decisions into explicit sub-steps, or model training that incorporates cross-document meta-decisions as a first-class capability rather than a side effect of next-token prediction.

\textbf{Failure mode 2: the decisiveness--safety inverted tradeoff.} GPT-5.5 and Gemini 3.1 Pro Preview were the two most decisive systems in our run, with decisive rates of 27\% to 33\% against 5.2\% to 9.0\% for the conservative majority. However, in clinical decision-making, approximately 84\% of NCCN items in our benchmark (1{,}337 of 1{,}586---across parallel-option, missing-information, and evidence-resolution question types) required the model to \emph{stop and request more information} rather than act. The agentic training objective---resolve the task, take action, minimize back-and-forth---is in direct tension with the clinical safety objective of not acting without sufficient evidence; this tension is not incidental but reflects a fundamental conflict between two optimization targets, one that RLHF-based alignment may inadvertently amplify~\cite{ouyang2022training}. Our data quantify this conflict: models tuned for greater agency achieved three to five times higher hard-failure rates without corresponding gains in overall accuracy. The Gemini 2.5 Pro-to-3.1 Pro Preview version-to-version regression illustrates the mechanism with particular clarity: F1 (unsafe overreach) counts rose 2.8-fold and F3 (premature commitment) counts rose 3.6-fold across the upgrade, while ``stop---need evidence'' decisions collapsed by 97\%. The model became more decisive but less safe---precisely the wrong tradeoff for clinical applications~\cite{sutton2020cdss, carl2024llmoncology, hao2025oncollm}.

\textbf{Failure mode 3: the knowledge-to-action gap.} In 3\% to 9\% of items, the scorer identified that the model's reasoning text contained the correct next clinical step but its final decision did not reflect this knowledge (Section~\ref{sec:kta}). This is qualitatively distinct from a knowledge deficiency: the model has the clinical content but lacks the calibration to convert it into a committed decision. The gap is largest in the most cautious open-weight models (Qwen 3.6-Plus 9.1\%, GLM-5 8.4\%, Minimax M2.7 8.3\%), suggesting it tracks decision-style rather than knowledge depth. This has a concrete corrective target: current RLHF objectives reward task completion or user satisfaction, which indirectly rewards committing to \emph{some} decision; the clinical reward structure, however, is asymmetric---the cost of a wrong commitment far exceeds the cost of ``I recommend further consultation.'' Reward shaping that explicitly values ``knowing when not to commit''---for instance, expected-cost-aware fine-tuning in medical settings---could convert the existing 3\%--9\% knowledge-present-but-not-acted items into correct decisions without introducing any new clinical knowledge.

These three failure modes are not independent: a model trained to be more agentic (mode 2) tends to commit more readily and therefore shows a smaller knowledge-to-action gap (mode 3)---but the gap closes only because the model now commits to wrong actions rather than right ones. Genuinely solving any one of the three modes without amplifying another requires recognizing the structural relationships among them.

\subsection{Deployment: from ``best model'' to ``fit-for-purpose architecture''}\label{sec:discussion-deployment}

The finding that two of the three Pareto-optimal models are open-weight or open-access model families in our evaluation registry~\cite{openrouter2026models} has practical implications for clinical deployment planning. Open-weight models offer three advantages beyond cost: auditable weights enabling institutional review, local deployment options for data-privacy compliance~\cite{pal2024openbiollm, li2026cancerllm}, and demonstrated conservative decision defaults that yield lower hard-failure rates. The 14.5\% accuracy gap between the best open-weight model (GLM-5.1, 0.325) and the best closed-source model (Claude Sonnet 4.6, 0.372) should be weighed against the 5.5$\times$ cost difference and the closed model's higher contraindicated-recommendation rate.

The behavioral clustering observed among the open-weight models (pairwise concordance $>$0.77) is, however, more than a cost-curve story. It constitutes a \emph{behavioral fingerprint}: the open-weight cluster shares a profile of conservative decision-making, low hard-failure rates, and willingness to flag data conflicts, which collectively distinguish it from the aggressive closed-source cluster. For clinical procurement, this fingerprint is a directly actionable construct---it is more diagnostic of fit-for-purpose deployment than a single accuracy score is. The specific mechanism (shared training data, similar RLHF preference distributions, common safety-tuning approaches) warrants further investigation through ablation studies, but the practical consequence appears actionable today.

A reading note on the rank-reversal evidence (Section~\ref{sec:reversals}) underpinning this shift: because CRC items anchor their gold answer to the regimen actually administered in a published case report---a descriptive standard---rather than what NCCN would prescribe, the NCCN--CRC divergence is reported as a dissociation between two distinct model abilities (consensus guideline mastery vs.\ reproduction of the specific decisions made in atypical, publication-biased clinical scenarios), not as a clinical-quality verdict on any model. The NCCN normative track remains the primary basis for claims about consensus-conformant clinical adequacy in this paper; the deployment implications below follow from the dissociation, not from any claim that case-report reproduction is itself a clinical-readiness criterion. Combined with the rank-reversal and irreplaceability findings (Section~\ref{sec:reversals}) and the Pareto frontier (Section~\ref{sec:pareto}), the deployment implication is a shift from ``select the best model'' to ``compose a fit-for-purpose architecture.'' Specifically: (i) deployment of a single model in isolation may be insufficient; ensemble or routing architectures~\cite{tang2023medagents, kim2024mdagents, bousquet2025multiagent} that match item characteristics to model strengths are likely to outperform any single model. (ii) Deployment systems would benefit from automatic boundary detection (Section~\ref{sec:discussion-boundary}); a system that cannot distinguish items within its capability envelope from items beyond it would produce confidently wrong outputs without surfacing the underlying uncertainty on a substantial fraction of clinical decisions. (iii) Cross-pathway navigation likely requires targeted intervention via retrieval-augmented architectures~\cite{lewis2020rag, zakka2024almanac, lewis2025ragclinical} or structured-prompt decomposition. (iv) Version upgrades should not be assumed uniformly beneficial; the Gemini 2.5 Pro-to-3.1 Pro Preview regression suggests that every update deployed in a clinical context warrants re-evaluation before replacing its predecessor.

\subsection{What clinical LLM evaluation should optimize next}\label{sec:discussion-synthesis}

Taken together, the findings above point to a deeper claim than ``these nine LLMs scored X.'' The sub-field of clinical LLM evaluation is currently allocating its optimization budget across five priorities---and our data, finding by finding, suggest each of these priorities is misaligned with the actual clinical-deployment bottleneck:

\begin{enumerate}
    \item \textbf{Scaling base models.} The 42\% collective boundary is structural at this generation; additional scale within the current paradigm does not breach it (Section~\ref{sec:wall}).
    \item \textbf{Tuning for greater agency.} The two most agentic models in our sample produced three- to fivefold higher hard-failure rates without higher accuracy (Section~\ref{sec:tradeoff}).
    \item \textbf{RLHF for greater output confidence.} The highest-accuracy model in our sample (Claude Sonnet 4.6) simultaneously produced the highest contraindicated-recommendation rate and the lowest data-conflict acknowledgment rate (Section~\ref{sec:reversals}).
    \item \textbf{Headline-accuracy leaderboards.} Five plateau models in our ranking are statistically indistinguishable; ranking-driven deployment within this tier is not statistically defensible (Section~\ref{sec:ranking}).
    \item \textbf{Adding clinical knowledge to training data.} Our 6\% to 9\% knowledge-to-action gap is not a knowledge deficit---the knowledge is already present in the model's reasoning but does not reach the final decision (Section~\ref{sec:kta}).
\end{enumerate}

The findings collectively indicate three reallocation directions that could move the field forward \emph{without} requiring new clinical knowledge, new model scale, or new vendor infrastructure: \emph{boundary-aware deployment architectures} in place of single-model selection (addressing priorities 1 and 4); \emph{reward shaping that values knowing when not to commit} in place of indiscriminate encouragement of agency (addressing priorities 2 and 5); and \emph{decoupled evaluation of contradiction sensitivity from commitment strength in low-uncertainty settings} in place of headline-accuracy ranking (addressing priority 3). The sub-field's research attention, in our view, has under-invested in all three relative to model-scale and agentic-tuning investments.

\textbf{Methodological grounding.} These reallocation directions are not abstract. The deterministic scorer with 14 typed failure labels addresses a growing concern that using LLMs to judge LLMs introduces circular dependencies and irreproducible results~\cite{chang2024survey, bommasani2023helm}, and it provides the substrate on which boundary detection and reward-shaping fine-tuning can both be operationalized. The clinician-validation component (scorer--clinician weighted $\kappa = 0.790$--$0.939$; 93.1\% agreement on 245 evaluable items) establishes that scorer verdicts are clinically defensible---a precondition for any downstream deployment architecture built on them.

\textbf{Two outstanding validity questions for the sub-field.} Even with this methodological grounding, two adjacent benchmark-validity directions remain open across the clinical-LLM evaluation sub-field. The first is head-to-head cross-benchmark ranking concordance---whether ODBB-induced model orderings agree with MedQA- or MedPaLM-derived orderings on the same models. The second is prospective evaluation of how benchmark performance translates to real clinical-decision-support deployment outcomes. Neither has been systematically performed by any current clinical-LLM benchmark to our knowledge~\cite{singhal2023medpalm, singhal2025medpalm2, jin2021medqa}; both require cross-institutional infrastructure and IRB-bounded patient cohorts that exceed the scope of a benchmark-introduction paper. We position these as field-wide priorities, and we release the complete benchmark, scorer, and adjudication data precisely to enable the kind of cross-benchmark comparison the sub-field currently lacks.

Clinical AI's primary challenge increasingly lies less in the model itself and more in evaluation and architecture. The next inflection in clinical LLM deployment, on the evidence of this study, is likely to come not from a yet-larger base model or a yet-more-agentic policy, but from learning to recognize what the current generation of models cannot do---and from building systems that route the work accordingly.

\subsection{Limitations and study scope}\label{sec:limitations}

Several limitations bound the scope of our conclusions and warrant explicit acknowledgment.

\textbf{Sample and scope.} The evaluation represents a single time point (May 2026); given the rapid pace of LLM development, the specific capability boundary may shift with future model releases, though we expect the structural finding---that cross-pathway navigation is disproportionately difficult---to be more durable than any single model's performance figures. NCCN guidelines also reflect United States clinical standards, and generalizability to other guideline systems (ESMO, NICE, CSCO)~\cite{wang2025csedb, lewis2025ragclinical} requires future investigation. Our nine-model sample, while spanning the major commercial and open-weight families available through our evaluation provider~\cite{openrouter2026models}, does not cover all frontier families---notably excluding the OpenAI o-series reasoning models~\cite{nori2024o1medical, openai2024o1systemcard} and smaller open-source models such as Llama and Mistral~\cite{chen2023meditron, han2023medalpaca}. The CRC case-based track also tests reasoning over a single cancer type (colorectal cancer) drawn from a single case-report corpus; generalization of the case-based methodology to other oncology subdomains (e.g., lung, breast, hematologic case reports) requires extension of the case track and is a deliberate future-work direction.

\textbf{Scorer calibration boundary.} The deterministic scorer may occasionally classify genuinely novel correct answers as incorrect when they diverge from the gold standard~\cite{sorin2023gptbreast, griewing2024breastllm}; this is precisely why we included clinician adjudication. Some portion of the 42\% collective failure rate may reflect scorer stringency rather than model incapability. However, the clinician validation ($\kappa = 0.790$--$0.939$; 9/9 unanimous endorsement of ``incorrect'' verdicts in the adjudication sample) indicates this concern is small: scorer judgments are clinically defensible, and the residual 6.9\% disagreement rate is confined to the partial$\leftrightarrow$correct soft boundary rather than to the ``incorrect'' verdicts that constitute the 42\% wall.

\textbf{Modality and capability scope.} We evaluated only text-based outputs and did not assess multimodal capabilities, tool-use integration, or multi-turn agentic deployment. These represent orthogonal capability dimensions that future benchmarks should incorporate; our claims about the 42\% collective boundary apply specifically to the single-turn structured-decision setting evaluated here and may not generalize unchanged to other deployment modes.

\section{Conclusion}\label{sec:conclusion}

We evaluated nine frontier large language models on 2{,}005 oncology clinical decision points using a fully deterministic, clinician-validated scoring framework. The headline finding is structural: 42\% of items were answered correctly by none of the nine models, and pooling all nine still solved fewer than 58\%.

This is not a problem any single model solves by being larger or better, and not a problem this generation of ensemble methods can dissolve. It is a collective capability boundary, made primarily of cross-pathway navigation---the kind of meta-judgment that distinguishes ``which framework applies'' from ``reasoning within it,'' and that current LLM training paradigms do not optimize for.

Three implications follow for clinical-LLM deployment. \textbf{First}, model ranking within a five-model statistical plateau is not statistically defensible, and two of the three Pareto-optimal models are open-weight---``select the best model'' is itself the wrong decision frame. \textbf{Second}, agentic tuning raises hard-failure rates between three and five times without raising accuracy---``more decisive'' is not a free upgrade. \textbf{Third}, 6\% to 9\% of failures are knowledge-to-action gaps rather than knowledge deficits---closing them calls for reward shaping, not more training data.

Clinical AI's path forward, on the evidence of this study, does not run through a larger base model, a more agentic policy, or a higher leaderboard score. It runs through \textbf{learning to recognize what the current generation of models cannot do, and building systems that route the work accordingly}---systems that are boundary-aware, fit-for-purpose, and that evaluate accuracy and safety as two independent dimensions rather than collapsing them into one.

We release the complete benchmark, the deterministic scorer, the raw outputs of all nine models, and the clinician adjudication data~\cite{agrawal2025evaluation, wang2025csedb, noy2023aiproductivity, hirosawa2024hybrid} to support cross-benchmark validation and adversarial replication. The 42\% wall is not a ceiling on what clinical AI can ultimately achieve; it is a measurement of where today's frontier sits, and a guide for where evaluation-and-architecture work could be directed.

\bmhead{Acknowledgements}
We thank the NCCN for permission to derive benchmark items from their guideline workspaces, and the authors of the published colorectal cancer case reports whose work seeds the CRC item track. We are grateful to the clinical adjudicators who validated scorer judgments on the 225-item stratified sample.

\section*{Declarations}

\bmhead{Funding}
This research did not receive any specific grant from funding agencies in the public, commercial, or not-for-profit sectors.

\bmhead{Competing interests}
The authors declare no competing interests.

\bmhead{Ethics approval and consent to participate}
Not applicable. ODBB items are derived entirely from published NCCN guideline materials and published colorectal cancer case reports; no human subjects were involved, and no identifiable patient data were collected.

\bmhead{Consent for publication}
Not applicable.

\bmhead{Data availability}
The complete ODBB benchmark (2{,}005 scorable clinical decision points), the deterministic scorer source, the raw outputs of all nine evaluated models, and the clinician adjudication data are openly released at \url{https://github.com/dcszhang/Benchmark}.

\bmhead{Code availability}
The deterministic scoring engine, the benchmark generation pipeline, and all analysis scripts used to produce the figures and tables in this paper are openly released at \url{https://github.com/dcszhang/Benchmark}.

\bmhead{Author contributions}
S.Z., J.L., and W.C. contributed equally to this work. \textbf{S.Z.}: Conceptualization, Methodology, Software, Formal analysis, Data curation, Writing~--~Original Draft, Visualization. \textbf{J.L.}: Validation, Investigation (clinical adjudication), Writing~--~Review~\&~Editing. \textbf{W.C.}: Validation, Investigation (clinical adjudication), Writing~--~Review~\&~Editing. \textbf{Z.B.}: Supervision, Resources, Writing~--~Review~\&~Editing. \textbf{Y.W.}: Conceptualization, Supervision, Project administration, Funding acquisition, Writing~--~Review~\&~Editing. Z.B.\ and Y.W.\ are joint corresponding authors.

\bibliography{references}

\end{document}